\documentclass[sigconf]{acmart}

\usepackage{epsfig,endnotes}
\usepackage{amssymb}
\usepackage{balance}
\usepackage{amsmath}
\usepackage{nicefrac}
\usepackage{siunitx}
\usepackage{array,framed}
\usepackage{booktabs}
\usepackage{
	color,
	float,
	epsfig,
	wrapfig,
	graphics,
	graphicx,
	subcaption
}
\usepackage{textcomp}
\usepackage{setspace}
\usepackage{latexsym,fancyhdr,url}
\usepackage{enumerate}
\usepackage[noend,ruled,linesnumbered]{algorithm2e}
\usepackage{algpseudocode}

\usepackage{graphics}
\usepackage{xparse} 
\usepackage{xspace}
\usepackage{multirow}
\usepackage{csvsimple}
\usepackage{balance}
\usepackage{threeparttable}

\usepackage{
	tikz,
	pgfplots,
	pgfplotstable
}
\usepackage{hyperref}
\usepackage{mathtools}
\usepackage{fancyhdr}
\usepackage[table]{xcolor}

\newtheorem{theorem}{Theorem}[section]  
\newtheorem{definition}[theorem]{Definition} 

\usepackage{amsmath}
\usepackage{xcolor}

\newcommand{\YH}[1]{\textcolor{blue}{[YH: ~#1]}}
\newcommand{\WW}[1]{\textcolor{red}{[WW: ~#1]}}
\newcommand{\Jeff}[1]{\textcolor{orange}{[Jeff: ~#1]}}

\def\g{{\bf g}}

\def\v{{\bf v}}

\newcommand{\system}{\textsf{PGR}}
\newcommand{\oldgraph}{G}
\newcommand{\atttarget}{G_{T}}
\newcommand{\newgraph}{\hat{G}}
\newcommand{\attackgraph}{G_{A}}
\newcommand{\oldmodel}{f}
\newcommand{\newmodel}{\hat{f}}

\newcommand{\nop}[1]{}

\newenvironment{ditemize}{%
\begin{list}{{\bf $\bullet$}}{%
\setlength{\itemsep}{0pt}\setlength{\rightmargin}{0pt}%
\setlength{\leftmargin}{1.2em}\setlength{\parsep}{0pt}}}{
\end{list}}

\copyrightyear{2025}
\acmYear{2025}
\setcopyright{acmlicensed}\acmConference[CCS '25]{Proceedings of the 2025
ACM SIGSAC Conference on Computer and Communications Security}{October
13--17, 2025}{Taipei, Taiwan}
\acmBooktitle{Proceedings of the 2025 ACM SIGSAC Conference on Computer
and Communications Security (CCS '25), October 13--17, 2025, Taipei, Taiwan}
\acmDOI{10.1145/3719027.3765173}
\acmISBN{979-8-4007-1525-9/2025/10}

\begin{document}

\title{Safeguarding Graph Neural Networks against Topology Inference Attacks}

\author{Jie Fu}
\affiliation{
  \institution{Stevens Institute of Technology}
    \country{Hoboken, NJ, USA}
}
\email{jfu13@stevens.edu}

\author{Yuan Hong}
\affiliation{%
  \institution{University of Connecticut}
  \country{Storrs, Connecticut, USA}
}
\email{yuan.hong@uconn.edu}

\author{Zhili Chen}
\affiliation{%
  \institution{East China Normal University}
  \country{Shanghai, China}
}
\email{zhlchen@sei.ecnu.edu.cn}

\author{Wendy Hui Wang}
\affiliation{%
  \institution{Stevens Institute of Technology}
  \country{Hoboken, NJ, USA}
}
\email{hwang4@stevens.edu}

\begin{abstract} 
Graph Neural Networks (GNNs) have emerged as powerful models for learning from graph-structured data. However, their widespread adoption has raised serious privacy concerns. While prior research has primarily focused on edge-level privacy, a critical yet underexplored threat lies in {\em topology privacy} —  the confidentiality of the graph’s overall structure. 
In this work, we present a comprehensive study on topology privacy risks in GNNs, revealing their vulnerability to graph-level inference attacks. To this end, we propose a suite of {\em Topology Inference Attacks} (TIAs) that can reconstruct the structure of a target training graph using only black-box access to a GNN model. Our findings show that GNNs are highly susceptible to these attacks, and that existing edge-level differential privacy mechanisms are insufficient as they either fail to mitigate the risk or severely compromise model accuracy. 
To address this challenge, we introduce {\em Private Graph Reconstruction}  (\system), a novel defense framework designed to protect topology privacy while maintaining model accuracy. \system\ is formulated as a bi-level optimization problem, where a synthetic training graph is iteratively generated using meta-gradients, and the GNN model is concurrently updated based on the evolving graph. 
Extensive experiments demonstrate that \system\ significantly reduces topology leakage with minimal impact on model accuracy. Our code is available at \url{https://github.com/JeffffffFu/PGR}.

\end{abstract}




\maketitle

\section{Introduction}
\label{sec:intro}




Graph data is pervasive in real-world applications, ranging from social networks~\cite{guo2020deep} to bioinformatics~\cite{sun2022does} and recommendation systems~\cite{huang2021knowledge}.
Graph Neural Networks (GNNs)~\cite{hamilton2017inductive,kipf2016semi} have recently demonstrated remarkable effectiveness in learning from such data. 

As GNNs are increasingly applied in various domains and applications, 
concerns have arisen regarding the data privacy risks associated with GNNs. Among these concerns, edge-level privacy has gained significant attention.
Recent studies have demonstrated that the adversary can infer the presence of specific edges in a graph by simply querying a GNN trained on the graph and analyzing its outputs~\cite{he2024maui, he2021stealing, wu2022linkteller, meng2023devil, wang2023link, zhong2023disparate}. Several privacy-preserving mechanisms (e.g.,~\cite{tang2024edge,kolluri2022lpgnet,yuan2023privgraph,sajadmanesh2023gap,wu2022linkteller}) have been designed to provide edge-level privacy protection with the differential privacy (DP) guarantee \cite{dwork2006calibrating,dwork2014algorithmic}. 


While extensive research has primarily focused on edge-level privacy in GNNs, {\em topology-level privacy} has been largely overlooked. Topology privacy refers to protecting the overall structure of the graph, including both the nodes and the connections between them in the training graph of the target model. Since the graph's structure encodes critical information about the relationships and dependencies among its nodes, revealing the graph's topology goes far beyond simply exposing a few edges. Inferring the graph topology can uncover hidden patterns and relationships within the network, which may be far more sensitive than knowledge of individual connections. For instance, in a social network graph, inferring the topology could enable an adversary to identify vulnerable or isolated communities and strategically target key individuals, such as influential users or celebrity figures, within those communities. Given the significance of topology privacy, it is crucial to investigate the vulnerabilities in GNNs regarding topology leakage and explore potential countermeasures. 

In this paper, we take the first step toward studying the topology privacy of GNNs. We conduct a comprehensive investigation into the topology privacy risks in GNNs, uncovering their vulnerability to graph-level inference attacks. Furthermore, we propose effective countermeasures to protect the topology privacy of GNNs. We make the following contributions. 

{\bf Attack design and evaluation.} We propose three Topology Inference Attacks (TIAs) - {\em metric-based TIA (M-TIA)}, {\em classifier-based TIA (C-TIA)}, and {\em influence-based TIA (I-TIA)}  - to reconstruct the overall structure of a private graph through black-box access to GNNs. To quantify the extent of topology privacy leakage (TPL), we define a metric based on the Jaccard similarity between the private graph and the graph inferred by a TIA. Our empirical results demonstrate the effectiveness of these attacks, with I-TIA achieving 100\% TPL, indicating complete leakage of the original graph’s topology. We further evaluate six state-of-the-art edge-level differential privacy (edge-DP) methods~\cite{tang2024edge,kolluri2022lpgnet,yuan2023privgraph,sajadmanesh2023gap,wu2022linkteller} against the proposed TIAs. The results reveal that none of these methods provide sufficient defense: they either fail to substantially reduce TPL or severely degrade model accuracy, leaving GNNs vulnerable to topology privacy attacks.

{\bf Design of \system.} We propose a novel algorithm, {\em Private Graph Reconstruction (\system)}, to mitigate topology privacy leakage in GNNs while preserving model performance. \system\ is formulated as a bi-level optimization problem that aims to generate a synthesized graph $\newgraph$ which contains no edges from the private training graph $\oldgraph$, yet enables any GNN trained on $\newgraph$ to achieve performance comparable to one trained on $\oldgraph$. To solve this optimization problem, \system\ iteratively constructs $\newgraph$ using meta-gradients while simultaneously updating the GNN on the synthesized graph. Furthermore, we introduce {\em DP-PGR}, an extension of \system\ with edge-level differential privacy guarantees, thereby providing protection at both the topology and edge levels.

{\bf Evaluation.} 
We conduct extensive experiments on four real-world graph datasets and three widely-used GNN models to evaluate the performance of \system. The results demonstrate that \system\ is highly effective in protecting topology privacy, significantly reducing topology privacy leakage. Moreover, \system\ consistently outperforms six state-of-the-art edge-DP approaches~\cite{tang2024edge,kolluri2022lpgnet,yuan2023privgraph,sajadmanesh2023gap,wu2022linkteller} in balancing model accuracy and topology privacy. Notably, \system\ incurs only a negligible loss in model accuracy, with a maximum decrease of just 0.1\%. 
Furthermore, we evaluate the robustness of \system\ against adversaries who possess prior knowledge of its design and parameters. Our findings show only a marginal decline (at most 0.4\%) in the effectiveness of \system, further demonstrating  its resilience in protecting topology privacy.

In summary, we make the following contributions.
\begin{ditemize}
\item We demonstrate that even simple adaptations of existing edge-level inference attacks can lead to significant topology leakage, uncovering a critical and previously underexplored privacy vulnerability in GNNs. 

\item We empirically show that six state-of-the-art edge-level DP methods are insufficient for protecting topology privacy. 

\item We introduce \system, the first defense mechanism specifically designed to mitigate topology privacy leakage in GNNs while maintaining model accuracy.


\item We perform extensive empirical evaluations of \system, demonstrating its ability to effectively balance topology privacy and model utility, while also maintaining robustness against adversaries with advanced knowledge. 
\end{ditemize}

\nop{
In summary, we made the following contributions. \WW{I don't think we need to repeat these contributions.}

\begin{itemize}

         \item We designed a measurement framework of topology privacy vulnerability. 
        This framework can perform an effective topology privacy audit on the private synthetic graph method under the black box model. 

        \item We proposed the \system~algorithm, which provides strong topology privacy protection while maintaining model utility.

        \item We introduce  DP-\system, demonstrating that the \system~algorithm can be efficiently integrated with differential privacy mechanisms to further enhance topology privacy protection.

        \item Extensive experiments validate the effectiveness of both the \system~in terms of model utility and privacy protection.
\end{itemize}

The rest of the paper is organized as follows. Section~\ref{sec:prelim} introduces the preliminary knowledge. Section~\ref{sec:problem} shows the problem setting and threat model. Topology privacy measurement framework is conducted in Section~\ref{sec:TPL} and our method \system\  are presented in Section~\ref{sec:methodology}.
In Section~\ref{sec:evalution}, we present the experiment evaluation of our method.  Section~\ref{sec:related} introduces related work, followed the conclusion in Section~\ref{sec:conclusion}. 

\YH{Section 1 is well written (only need minor edits in the final pass)}
}  

\section{Related Work}
\label{sec:related}

\subsection{Privacy Inference Attacks against GNNs} 

Recent studies have revealed that GNNs are vulnerable to the membership inference attacks (MIAs)~\cite{duddu2020quantifying,olatunji2021membership,wang2024subgraph}. 
Most existing MIAs against GNNs focus on edge inference. Specifically, a {\em link membership inference attack} (LMIA) seeks to infer the presence of specific edges in the training graph of a GNN. 
LMIAs can be broadly classified into two categories: (i) {\em similarity-based LMIAs}~\cite{he2021stealing,wang2023link,wang2024gcl}, which leverage the intuition that nodes with higher similarity are more likely to be connected, and (ii) {\em influence-based LMIAs}~\cite{wu2022linkteller,meng2023devil,he2024maui}, which use the influence between node pairs for inference. Influence is estimated by applying perturbations to either the node features~\cite{wu2022linkteller,he2024maui} or the node's neighborhood~\cite{meng2023devil}. A higher influence between two nodes indicates a greater likelihood that an edge exists between them.  We will demonstrate that even a simple adaptation of LMIAs can  compromise topology privacy (Section~\ref{sc:attack}).

More recently, a new class of attacks, referred to as {\em graph reconstruction attacks}, has emerged. These attacks aim to reconstruct the training graph~\cite{wang2024subgraph,zhang2022inference,duddu2020quantifying} through access to GNN models. Wang et al.~\cite{wang2024subgraph} introduced a black-box attack, SMIA, designed to recover specific graph structures (e.g., $k$-hop paths or cliques) from the training graph, differing from TIA in its attack goal. Zhang et al.~\cite{zhang2022inference} proposed a white-box attack that reconstructs a graph with similar graph statistics (e.g., degree distribution, local clustering coefficient) to the target graph by leveraging its embedding. Duddu et al.~\cite{duddu2020quantifying} also developed a white-box attack that reconstructs the graph using an encoder-decoder model on graph embeddings. However, TIA differs fundamentally from these attacks as it only requires access to the posterior probabilities output by the model, rather than node embeddings~\cite{duddu2020quantifying} or graph embeddings~\cite{zhang2022inference}.



\vspace{-0.1in}

\subsection{Differentially Private GNNs}


Differential privacy (DP)~\cite{dwork2006calibrating} is a robust mathematical framework that formally defines the privacy guarantees an algorithm can provide. DP can be applied to graph data in two primary ways~\cite{sajadmanesh2023gap}: {\em edge-level DP} and {\em node-level DP}. In edge-level DP, two graphs are considered neighbors if they differ by a single edge, while node-level DP considers two graphs neighbors if they differ by the edges connected to a single node. Achieving node-level DP can be particularly challenging, as modifying one node may remove up to $K-1$ edges in the worst case, where $K$ is the number of edges in the graph. Therefore, this paper focuses solely on edge-level DP.

Recently, there have been efforts to provide formal edge-level DP privacy guarantees within various GNN learning settings. Wu et al.~\cite{wu2022linkteller} introduced methods such as {\em EdgeRand} and {\em LapEdge}, which perturb the input graph using randomized response or the Laplace mechanism, respectively. The perturbed adjacency matrix is then used for GNN training. Tang et al.~\cite{tang2024edge} proposed {\em Eclipse}, a method that transforms the adjacency matrix into a low-rank form using singular value decomposition (SVD) before adding Laplace noise to the singular values. Yuan et al.~\cite{yuan2023privgraph} developed {\em PrivGraph}, an edge-DP approach that partitions nodes into communities and adds noise to both intra-community and inter-community edges. The graph is then synthesized using the CL model~\cite{aiello2000random}. Sajadmanesh et al.~\cite{sajadmanesh2023gap} designed {\em GAP}, which achieves edge-DP by adding calibrated Gaussian noise to the output of the GNN's aggregation function. Kolluri et al.~\cite{kolluri2022lpgnet} introduced {\em LPGNet}, which perturbs cluster degree vectors with Laplace noise during training, using these noisy vectors as input features for model training. However, as we will show in Section~\ref{sc:attack}, these edge-DP methods fail to provide adequate protection for topology-level privacy. 

\subsection{Graph-level Privacy Protection} 

Most existing graph-level privacy protection methods~\cite{han2024subgraph,qiu2022privacy,zhang2023extracting} focus on the Federated Learning (FL) setting, aiming to protect the privacy of local graphs on the client side. Specifically, Han et al.~\cite{han2024subgraph} address privacy in recommendation systems, expanding local graphs leveraging Software Guard Extensions (SGX) and Local Differential Privacy (LDP). 
Qiu et al.~\cite{qiu2022privacy} propose {\em DP-FedRec}, a federated GNN that employs Private Set Intersection (PSI) to extend local graphs for each client, tackling the non-IID challenge in the FL setting. Zhang et al.~\cite{zhang2023extracting} take a subgraph-level approach, utilizing the {\em information bottleneck  principle} to extract task-relevant subgraphs from local client graphs for training and local model updates. While these methods address challenges like non-IID distributions and safe gradient sharing within the FL framework, their goals and strategies differ fundamentally from ours.

\section{Problem Formulation} \label{sec:problem}

\begin{table}[t!]
\small
    \centering
    \begin{tabular}{c|c}\hline
        Symbol& Meaning  \\\hline
         $\oldgraph$/$\newgraph$ & Original graph/synthetic graph\\\hline
         $\oldmodel$/$\newmodel$ & GNN trained on $\oldgraph$/$\newgraph$ \\\hline
         $\atttarget$/$\attackgraph$ & Target graph/Attack graph \\\hline
         $E$/$E_T$/$E_A$ & Edge set of $\oldgraph$/$\atttarget$/$G_A$ \\\hline
         $K$/$\hat{K}$ & \# of edges of $\oldgraph$/$\newgraph$\\\hline
         $K_T$/$K_A$& \# of edges in $\atttarget$/adversary knowledge\\\hline         
    \end{tabular}
    \caption{Common notations}
    \vspace{-0.2in}
    \label{tab:notation}
\end{table}

Given an undirected graph $G(V, E)$, where $V$ and $E$ denote the nodes and edges of $G$, respectively, a GNN model aims to learn a low-dimensional embedding for each node by leveraging both the node features and the overall topology of the graph. The learned node embeddings can 
then aid downstream graph learning tasks, such as node classification, link prediction, and clustering. In this paper, we consider node classification as the learning task. Specifically, for any given node $v\in G$, the GNN model outputs a {\em posterior probability vector} for $v$, in which the $i$-th entry indicates the probability that $v$ is associated with the $i$-th label. 
Specifically, we consider node classification in the transductive setting, given its wide use in practice. 
 Table \ref{tab:notation} lists the common notations used in the paper. 

\subsection{Threat Model}
\label{sc:threatmodel}

An attacker may seek to infer topological information about the training graph of a GNN model. This information could pertain to either the entire graph or a specific subgraph. For example, consider an attacker planning a cyberattack on a computer network. The attacker might target the entire network, attempting to reconstruct the topology of the full network graph, or focus on a particular subnet, aiming to infer the structure of the corresponding subgraph.

Therefore, given a graph $\oldgraph(V, E)$ and a GNN model $\oldmodel$ trained on $\oldgraph$, the adversary's goal is to infer the topology of a {\em target graph} $\atttarget(V_T, E_T)\subseteq\oldgraph$ where $V_T\subseteq V, E_T\subseteq E$. We assume the adversary has the following knowledge: 

\begin{ditemize}
\item {\bf Black-box model access.} Following the prior works  \cite{he2021stealing}, we assume the adversary has black-box access to $\oldmodel$. Specifically, the adversary can access $\oldmodel(v)$ for any query node $v$, where $\oldmodel(v)$ outputs a posterior probability vector. 
Several machine-learning-as-a-service providers, such as Google Cloud's Vertex AI~\cite{Vertex}, Facebook's ParlAI~\cite{Parlai}, and IBM's InfoSphere Virtual Data Pipeline~\cite{Infosphere}, offer such black-box access. Consistent with existing LMIAs \cite{he2021stealing,wu2022linkteller}, we also assume the adversary has an unlimited query budget. 

\vspace{0.05in}

\item {\bf Auxiliary graph information.} We consider an adversary who has knowledge of the nodes and their features in the training graph - and consequently in the target graph $\atttarget$ - as assumed in prior works \cite{he2021stealing,wu2022linkteller}. In addition, the adversary possesses the auxiliary information about the number of edges in $\atttarget$. This estimate is presumed to be close to the true value, i.e., $K_{A} \approx K_T$, where $K_T$ represents the actual number of edges in $\atttarget$. Such information is often readily available from external sources and may even be publicly accessible, even when the entire training graph is the attack target. For instance, data agencies like Statista~\cite{Statista} frequently publish statistics on the number of nodes and edges for major social networks, such as Facebook and Instagram. 
\end{ditemize}

\begin{figure}[t!]
	\begin{center}
  \includegraphics[width=0.7\linewidth]{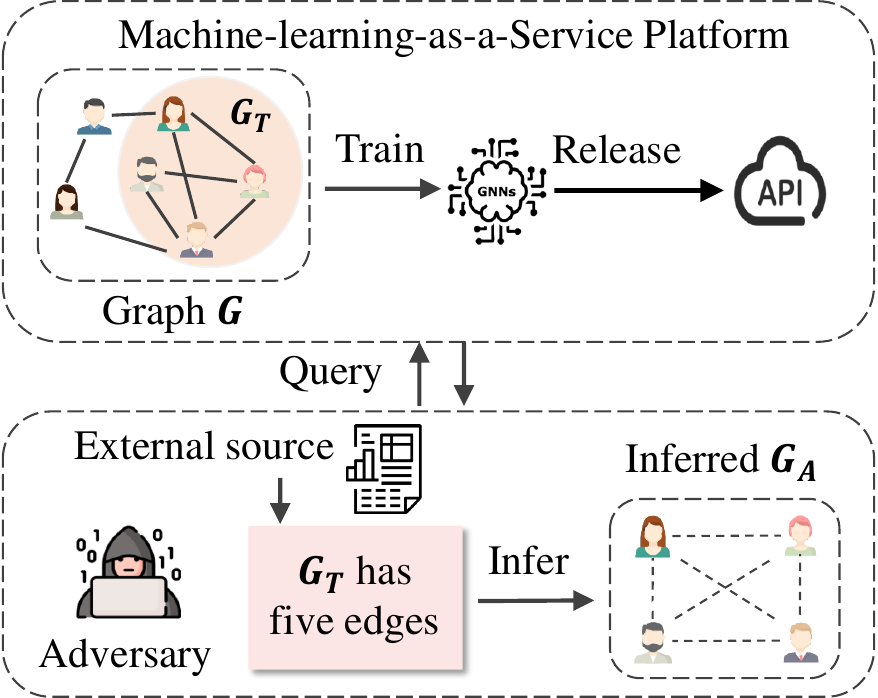}
		\caption{An example of the topology inference attack. The target graph $G_T$, which is a subgraph of $G$, is shaded. }\vspace{-0.1in}
        \label{figure:threat model}
	\end{center}
\end{figure}

Given these knowledge, the attacker's goal is to infer a graph $\attackgraph(V_A, E_A)$ where  $V_A=V_T$ and $|E_A|=K_A$. 
Figure \ref{figure:threat model} illustrates an example of the attack at a high level. 
Compared with existing LMIAs \cite{wu2022linkteller,he2021stealing}, our adversary has additional knowledge about the number of edges in the training graph. 
We will show that such a simple piece of adversary knowledge will bring the privacy risks to the graph  topology, revealing new privacy vulnerabilities in GNNs. 

As the graph structure carries a wealth of information about the relationships and dependencies between its nodes, revealing graph topology goes beyond simply exposing a few edges. The inference of graph topology can reveal hidden structures and relationships in the network, which can be far more sensitive than just knowing a few individual connections. 
For example, as the topology of a graph often reveals clusters or communities of closely connected nodes, it allows the adversary to identify the existence of these communities, leading to potential target attacks against specific user groups. Furthermore, deriving a node's degree  or its centrality in the network from the topology can help the attacker to target key players or potential weak points in the network.

\subsection{Measurement of Topology Privacy Leakage}

Measuring topology privacy leakage is crucial for understanding how much sensitive information about a network's structure might be exposed. By quantifying this leakage, we can assess the risk of revealing key characteristics such as node relationships and connectivity patterns within the network. In this paper, we consider similarity measures to help evaluate the extent of the training graph's topology being inferred by the attacks. \footnote{We also examined an alternative TPL metric and found its results to be consistent with those obtained using the Jaccard similarity metric. Further details are provided in Section~\ref{sc:discussion}.} 

Specifically, given a graph $\oldgraph(V, E)$ and a GNN model $\oldmodel$ trained on $\oldgraph$, let $\atttarget(V_T, E_T)$ be the private graph, where $V_T\subseteq V, E_T\subseteq E$. Let $\attackgraph$ be the {\em attack graph} inferred by the adversary through black-box access to $\oldmodel$, and $E_{A}$ be the edges of $\attackgraph$. We quantify the {\em topology privacy leakage} of $\oldmodel$ as the Jaccard similarity between $E_T$ and $E_{A}$: 

\begin{align} \label{equ:first part loss}
       \begin{split}
\begin{aligned}
TPL = \frac{|E_T \cap E_{A}|}{|E_T \cup E_{A}|}.
\end{aligned}
	\end{split}
\end{align}

The Jaccard similarity is directly related to True-Positive (TP), False-Positive (FP), and False-Negative (FN) edges, 
where TP edges are those that are present in both graphs, FP edges are those that appear in $E_{A}$ but not in $E_T$, and FN edges are those are missing in $E_{A}$ but are present in $E_T$. 
Specifically, the numerator $|E_T \cap E_{A}|$ corresponds to the TP edges, while the denominator corresponds to the total number of unique edges that appear in either $E_T$ or $E_{A}$, which is calculated as $|TP|+|FP|+|FN|$.  
Thus, the Jaccard similarity can be written in terms of TP and FP as: $TPL=\frac{|TP|}{|TP|+|FP|+|FN|}$. A higher Jaccard similarity indicates that the graphs are more similar, meaning there are more true-positive edges relative to false positives and false negatives. Conversely, a lower Jaccard similarity means the predicted graph has many false positives or false negatives, indicating less accuracy in representing the true graph's topology. $TPL$ reaches the maximum  (i.e., $TPL=1$) when $E_T=E_{A}$, and the minimum (i.e., $TPL=0$) when $E_T\cap E_{A}=\emptyset$.

\section{Topology Inference Attack}
\label{sc:attack} 


Given the adversary knowledge, the attacker performs a {\em topology inference attack} (TIA) to construct a graph whose topology is consistent with the known number of edges. Formally, given a training graph $\oldgraph(V, E)$, let $\oldmodel$ be the GNN model trained on $\oldgraph$, and $\atttarget(V_T, E_T)\subseteq\oldgraph$ be the target private graph. The adversary aims to construct an attack graph $\attackgraph(V_T, E_{A})$ where $|E_{A}|=K_{A}$, through black-box access to $\oldmodel$. 

Intuitively, the adversary can construct all possible graphs whose number of edges equals to $K_{A}$, and randomly pick one from these graphs as $\attackgraph$. However, such naive method is prohibitively costly due to its large search space.\footnote{Given $N$ nodes, there are $\binom{N(N-1)/2}{K}$ possible graphs that have the same $N$ nodes and $K$ edges.} Furthermore, as it neglects the output of the GNN model, its accuracy is poor (as shown in Appendix \ref{Appendix:randomguess}). Therefore, we aim to design more effective TIAs through utilizing the black-box access of $\oldmodel$. 

Intuitively, the graph topology can be inferred by reconstructing the individual edges. Building upon this intuition, we extend existing LMIAs \cite{wu2022linkteller, he2021stealing, meng2023devil} to TIAs, broadening the inference from a single edge to a set of edges.  
Our goal is to show that topology inference does not require the design of entirely new attack strategies - rather, even straightforward adaptations of existing LMIAs can pose significant threats to topology privacy. 

\subsection{Attack Details}

Based on the underlying reasoning of the attacks, we classify our TIAs into two types: {\em similarity-based} and {\em influence-based}. Next, we present the details of each type of attacks.

\noindent{\bf Similarity-based Topology Inference Attack.} In this attack, the adversary constructs edges in $\attackgraph$ based on the similarity between the posterior probability of nodes. Specifically, we design two TIAs that exploit this similarity to infer the graph structure:

\begin{ditemize}
    \item {\bf Metric-based TIA (M-TIA)}: The adversary computes the similarity between the posterior probability vectors of each node pair $(u_i, v_j)$ in graph $G$ using a distance-based metric.\footnote{We consider three distance functions: Cosine distance, Chebyshev distance, and Euclidean distance. In our experiments, we report the highest attack accuracy achieved across these three metrics.} The top $K_{A}$ node pairs with the highest similarity scores are then selected, and edges are inserted between them in $\attackgraph$.
    
    \vspace{0.05in}
    
    \item {\bf Classifier-based TIA (C-TIA)}: The adversary leverages an existing LMIA, such as~\cite{he2021stealing,wang2024gcl}, to construct $\attackgraph$. Specifically, for each node pair $(v_i, v_j)$, the adversary queries the LMIA classifier to obtain both a membership prediction label and a corresponding confidence score. Among the node pairs predicted as members (i.e., links), those with the top $K_{A}$ highest confidence scores are selected and inserted as edges in $\attackgraph$.
\end{ditemize}


\smallskip
\noindent{\bf Influence-based Topology Inference Attack (I-TIA).} 
We adapt the influence-based LMIAs~\cite{wu2022linkteller, meng2023devil} to the topology inference setting. Specifically, for each node pair $(v_i, v_j)$, we compute the influence of node $v_i$ on node $v_j$. In this work, we adopt the pairwise influence score function proposed in~\cite{meng2023devil}. The $K_{A}$ node pairs with the highest influence scores are then selected and added as edges in $\attackgraph$. 

While our TIAs are inspired by existing edge-level inference attacks, they represent a significant advancement by moving beyond isolated link prediction to reconstruct broader topological structures. As we will demonstrate in Section~\ref{sc:eval-edge-dp}, even such simple adaptations of edge-level attacks can lead to significant topology leakage, highlighting a critical and previously underexplored privacy vulnerability in GNNs.

\subsection{Insufficiency of Edge-DP as Defense}
\label{sc:eval-edge-dp}

There have been active attempts to provide edge-level differential privacy (edge-DP) in GNNs. This raises the following concern: {\em While these approaches can provide rigorous privacy protection at the edge level, are they able to ensure privacy at the topology level? Specifically, what will be the privacy leakage of these approaches against TIAs?} We will answer this question through empirical evaluations. 

\begin{table*}[!t]
\caption{Topology privacy leakage (\%) and node classification accuracy (Utility) of GCN model with and without edge-DP protection. The privacy budget of all edge-DP approaches is set as $\epsilon=7$. }
\small
\centering
\begin{tabular}{c|cccc|cccc|cccc}
\hline
\multirow{2}{*}{\textbf{Approach}} & \multicolumn{4}{c|}{\textbf{Cora}} & \multicolumn{4}{c|}{\textbf{LastFM}} & \multicolumn{4}{c}{\textbf{Emory}} \\ \cline{2-13} 
                          & \multicolumn{1}{c|}{\textbf{M-TIA}} & \multicolumn{1}{c|}{\textbf{C-TIA}} & \multicolumn{1}{c|}{\textbf{I-TIA}} & \textbf{Utility} & \multicolumn{1}{c|}{\textbf{M-TIA}} & \multicolumn{1}{c|}{\textbf{C-TIA}} & \multicolumn{1}{c|}{\textbf{I-TIA}} & \textbf{Utility} & \multicolumn{1}{c|}{\textbf{M-TIA}} & \multicolumn{1}{c|}{\textbf{C-TIA}} & \multicolumn{1}{c|}{\textbf{I-TIA}} & \textbf{Utility} \\ \hline
No Edge-DP          & \multicolumn{1}{c|}{28.10} & \multicolumn{1}{c|}{25.53} & \multicolumn{1}{c|}{100.00}   & 83.38   & \multicolumn{1}{c|}{25.04} & \multicolumn{1}{c|}{25.11} & \multicolumn{1}{c|}{100.00}   & 81.75   & \multicolumn{1}{c|}{17.55} & \multicolumn{1}{c|}{16.54} & \multicolumn{1}{c|}{100.00}   & 90.68   \\ \hline
Eclipse~\cite{tang2024edge}        & \multicolumn{1}{c|}{18.58} & \multicolumn{1}{c|}{16.46} & \multicolumn{1}{c|}{29.50}  & 72.45   & \multicolumn{1}{c|}{16.48} & \multicolumn{1}{c|}{5.93}  & \multicolumn{1}{c|}{20.94} & 73.87   & \multicolumn{1}{c|}{20.07} & \multicolumn{1}{c|}{15.91} & \multicolumn{1}{c|}{27.36} & 83.78   \\ \hline
LPGNet~\cite{kolluri2022lpgnet}     & \multicolumn{1}{c|}{24.67} & \multicolumn{1}{c|}{24.23} & \multicolumn{1}{c|}{17.12} & 70.13   & \multicolumn{1}{c|}{19.67} & \multicolumn{1}{c|}{22.74} & \multicolumn{1}{c|}{13.76} & 56.82   & \multicolumn{1}{c|}{18.96} & \multicolumn{1}{c|}{19.38} & \multicolumn{1}{c|}{17.16} & 71.38   \\ \hline
PrivGraph~\cite{yuan2023privgraph}  & \multicolumn{1}{c|}{2.94} & \multicolumn{1}{c|}{1.89}  & \multicolumn{1}{c|}{13.38} & 43.88   & \multicolumn{1}{c|}{19.21} & \multicolumn{1}{c|}{18.34} & \multicolumn{1}{c|}{13.87} & 46.89   & \multicolumn{1}{c|}{17.06} & \multicolumn{1}{c|}{11.56} & \multicolumn{1}{c|}{14.74} & 50.43   \\ \hline
GAP~\cite{sajadmanesh2023gap}      & \multicolumn{1}{c|}{4.63} & \multicolumn{1}{c|}{3.86}  & \multicolumn{1}{c|}{1.47}  & 59.45   & \multicolumn{1}{c|}{14.63} & \multicolumn{1}{c|}{16.53} & \multicolumn{1}{c|}{4.25}  & 67.55   & \multicolumn{1}{c|}{13.89} & \multicolumn{1}{c|}{11.54} & \multicolumn{1}{c|}{12.06} & 67.97   \\ \hline
LapEdge~\cite{wu2022linkteller}    & \multicolumn{1}{c|}{26.11} & \multicolumn{1}{c|}{25.02} & \multicolumn{1}{c|}{72.24} & 61.63   & \multicolumn{1}{c|}{25.01} & \multicolumn{1}{c|}{25.01} & \multicolumn{1}{c|}{58.12} & 54.89   & \multicolumn{1}{c|}{18.29} & \multicolumn{1}{c|}{15.40}  & \multicolumn{1}{c|}{95.30}  & 87.10   \\ \hline
EdgeRand~\cite{wu2022linkteller}   & \multicolumn{1}{c|}{25.52} & \multicolumn{1}{c|}{25.08} & \multicolumn{1}{c|}{96.22} & 74.04   & \multicolumn{1}{c|}{19.63} & \multicolumn{1}{c|}{18.41} & \multicolumn{1}{c|}{98.66} & 69.75   & \multicolumn{1}{c|}{17.52} & \multicolumn{1}{c|}{13.89} & \multicolumn{1}{c|}{98.98} & 89.05   \\ \hline
\end{tabular}
\label{table: evaluation of topology privacy loss}
\end{table*}

\smallskip
\noindent {\bf Experimental setup.} For this part of evaluation, we set up a two-layer Graph Convolutional Networks (GCN)~\cite{traud2012social} with a 32-neuron hidden layer as the target model. We utilize the ReLU activation function and Softmax function for the output.
We consider six state-of-the-art edge-DP approaches: {\em Eclipse}~\cite{tang2024edge}, {\em LPGNet}~\cite{kolluri2022lpgnet}, {\em PrivGraph}~\cite{yuan2023privgraph}, {\em GAP}~\cite{sajadmanesh2023gap}, {\em LapEdge}~\cite{wu2022linkteller}, and
{\em EdgeRand}~\cite{wu2022linkteller}. 

We conducted the experiments on four real-world graph datasets. \footnote{The descriptions and statistics of these datasets can be found in Table~\ref{table:Statistics of datasets}.} For each dataset, we sample five subgraphs.~\footnote{We randomly select a node from the graph as the starting point, and utilize the Breadth First Search (BFS) strategy to obtain edges in the sampled graph.} Each subgraph contains 100 nodes ($\sim$5\%-10\% of the original graph). 
We consider these sampled subgraphs as the target private graphs. 

Regarding the implementation of TIAs, we utilized {\em StealLink}~\cite{he2021stealing} for both M-TIA and C-TIA, and {\em Link-Infiltrator}~\cite{meng2023devil} for I-TIA. 
We do not use the same LMIA across three types of TIAs as StealLink is not suitable for I-TIA  as it does not rely on the node influence for privacy inference, while Link-Infiltrator cannot be used for both M-TIA and C-TIA as it does not utilize node similarities. In our evaluation, we assume that the adversary has knowledge of the exact number of edges in the  target private graph (i.e., $K_A = |E_T|$). This assumption allows for the strongest possible attack, which will serve as the primary basis for designing our defense strategy.

To assess the privacy risk of GNNs against TIAs, we measure the topology privacy leakage (Eqn. \eqref{equ:first part loss}) of the target GNN model with and without edge-DP protection. Furthermore, we measure {\em model utility} as the {\em accuracy of node classification}, i.e., the percentage of nodes whose labels are correctly predicted. 


\smallskip
\noindent{\bf Results.} Table~\ref{table: evaluation of topology privacy loss} presents the results of both topology privacy leakage (TPL) and model utility across three types of TIAs, with the privacy budget $\epsilon$ set to 7.\footnote{We chose $\epsilon = 7$ as it best strikes the balance between model accuracy and edge-level privacy. Additional results on the privacy-accuracy trade-off can be found in Appendix~\ref{appendix:C1}.} The TPL result is reported as the average TPL across the five subgraphs, whereas model utility is evaluated on the entire testing graph. 

First, we observe that in the absence of edge-DP protection, the GCN model is consistently vulnerable to all three TIAs, exhibiting substantial topology privacy leakage (TPL) across all settings. Among the three, the model is particularly susceptible to I-TIA, with TPL reaching as high as 100\%. This suggests that the node influence exploited by I-TIA more effectively captures structural information than M-TIA and C-TIA, likely due to its alignment with the message-passing mechanism inherent to GNNs.

Second, although all six edge-DP approaches demonstrate some level of protection against TIAs, their effectiveness varies significantly. In particular, LapEdge and EdgeRand provide the weakest protection, whereas GAP consistently offers the strongest defense, in most settings. The limited effectiveness of LapEdge and EdgeRand stems from their inability to alter the ranking of nodes’ similarity and influence - rankings that are computed from posterior probabilities. Although these probabilities are perturbed, the relative order often remains unchanged, resulting in reconstructed graphs that closely resemble the original despite the added noise. In contrast, GAP is capable of perturbing both the nodes' posterior probabilities and their induced rankings, thereby demonstrating strong effectiveness against TIA.

However, although certain edge-DP approaches - such as GAP and PrivGraph - are effective in significantly reducing TPL, they incur substantial degradation in model utility. For example, on the Cora dataset, GAP reduces the model utility from 83.38\% to 59.43\%, which is close to random guessing in a binary classification task.

In summary, none of the six edge-DP methods offers a satisfactory defense against TIAs. They either fail to significantly reduce the TPL or cause a substantial drop in model accuracy, making them ineffective as mitigation solutions. 

\section{\system: Our Countermeasure of TIA}
\label{sec:methodology}



\nop{
In general, there are two requirements for the design of a topology privacy protection algorithm: 
(i) The algorithm should minimize the topology privacy leakage (Eqn.~\eqref{equ:first part loss}) by TIAs; 
(ii) The algorithm should minimize the loss of model accuracy incurred by privacy protection. 
}

Intuitively, the minimal topology privacy leakage (Eqn.~\eqref{equ:first part loss}) occurs when there is no overlap between the edges of $\atttarget$ and $\attackgraph$. However, the defender typically lacks prior knowledge of the  target private graph $\atttarget$. Therefore, the defender must consider all possible subgraphs as potential attack targets. The most effective strategy to protect against all such possibilities is to construct a new graph $\newgraph$ whose edge set is entirely disjoint from that of the training graph $\oldgraph$. A new GNN model, $\newmodel$, is then trained on $\newgraph$. Since $\newgraph$ contains no edges from $\oldgraph$, the model is expected to reveal minimal structural information about both $\oldgraph$ and thus any potential target graph. 

Following this intuition, we design \system, a new privacy protection method to mitigate the TPL of GNN models.  
\system\ aims to serve a dual goal:

\begin{ditemize}
  \item {\bf Goal 1 (Privacy)}: Given a graph $\oldgraph(V, E)$, constructs a graph $\newgraph (V, \hat{E})$ such that $\hat{E} \cap E = \emptyset$. 

 \item {\bf Goal 2 (Model accuracy)}: The performance of the model $\newmodel$ trained on $\newgraph$ is close to that of $\oldmodel$ trained on $\oldgraph$.  
\end{ditemize}

\vspace{0.05in}


\begin{figure}[t!]
	\begin{center}
\includegraphics[width=\linewidth]{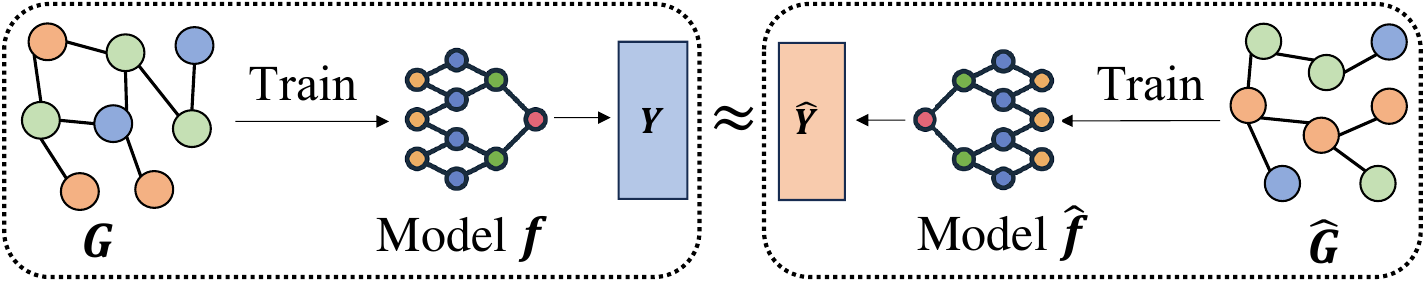}
 \vspace{-0.1in}
		\caption{Illustration of \system.}
        \label{figure:PGR_outline}
	\end{center}
    \vspace{-0.2in}
\end{figure}

Figure \ref{figure:PGR_outline} illustrates the framework of \system. Next, we discuss the details of \system. 




\subsection{Bi-level Optimization} \label{sec: Modeling the Problem}




First, we formalize the optimization problem of $\system$. 

\smallskip
\noindent{\bf Admissible space of graph construction.}  Intuitively, the topology privacy leakage is minimized if $\oldgraph$ and $\newgraph$ do not share any edge (i.e., $E\cap\hat{E}=\emptyset$). 
Therefore, we define the {\em admissible space}  (denoted as $\Phi(G)$) of $\newgraph$ w.r.t. a given graph $\oldgraph$ is thus defined as 
\[\Phi(G) =\{\hat{G}\ |\   E\cap\hat{E} =\emptyset\},\] 
where $E$ and $\hat{E}$ are the edge set of $\oldgraph$ and $\newgraph$, respectively. $\newgraph$ have the same sets of labeled nodes as $\oldmodel$. These nodes in $\newgraph$ are associated with the same labels as their corresponding counterparts in $\oldgraph$. The node features of $\newgraph$ remain the same as $\oldgraph$. 

\noindent{\bf Objective function.} Recall that one of the goals of \system\ is to ensure the performance of the model $\newmodel$ trained on $\newgraph$ is close to that of the model $\oldmodel$ trained on $\oldgraph$.  We formulate this as a bi-level optimization problem, with $\newgraph$ and $\theta$ as the outer-level and inner-level variables:

\begin{align} \label{equ:Bi-levl}
       \begin{split}
	\begin{aligned}
        \hat{G}^* &=  \overbrace{\underset{\newgraph\in\Phi(G)}{\arg \min } \ \mathcal{L}_{gen}\left(\hat{f}(\newgraph;\theta^*), {Y_P}\right)}^{\text{outer-level objective}} \\
     \text { s.t. } \quad \theta^* &=\underbrace{\underset{\theta}{\arg \min } \ \mathcal{L}_{train}\left(\hat{f}(\newgraph; \theta), Y_L\right)}_{\text{Inner-level objective}} 
		\end{aligned}
	\end{split}
\end{align}
where $\Phi(G)$ represents the admissible space of $\newgraph$,
$Y_L$ denotes the labels of the labeled nodes in $\oldgraph$, and $Y_P$ denotes the predicted labels of the unlabeled nodes in $\oldgraph$ predicted by $\oldmodel$. 
$\mathcal{L}_{gen}$ and $\mathcal{L}_{train}$ represent the loss function of node classification evaluated on $Y_P$ and $Y_L$, respectively. \footnote{We use cross-entropy for both $\mathcal{L}_{train}$ and $\mathcal{L}_{gen}$.} 
In other word, the inner-level objective is to train a model $\newmodel$ on the graph $\newgraph$, while the outer-level objective is to generate the graph $\newgraph$ such that the model $\newmodel$ trained on $\newgraph$ approximates the performance of the model $\oldmodel$ on  $\oldgraph$. 



\vspace{-0.1in}
\subsection{Meta-Gradient} \label{sec:Graph Reconstruction via Meta-Gradients}

Exactly solving Eqn.~\eqref{equ:Bi-levl} requires computing the gradient of the outer objective with respect to $\hat{G}$, which in turn involves fully solving the inner optimization problem to obtain the model parameters $\theta$ (i.e., until convergence). This process is computationally intensive. To address this, we propose an approximate solution based on the meta-gradient technique. Specifically, we treat the number of edges $\hat{K}$ in the reconstructed graph $\newgraph$ as a hyperparameter, and use the meta-gradient approach \cite{wan2021graph} to efficiently search for a graph with $\hat{K}$ edges that enables the GNN to achieve prediction accuracy comparable to that trained on the original graph.


Intuitively, meta-gradients approximate the optimal solution to a bi-level optimization problem by using truncated optimization — that is, updating the inner model parameters for only a few steps rather than running to full convergence — and then backpropagating through these intermediate updates. Specifically, we compute the meta-gradient $\nabla_{\newgraph}^{\text{meta}}$ of the outer loss $\mathcal{L}_{\text{gen}}$ with respect to the hyperparameter $\newgraph$: 
\begin{align} 
\label{eqn:meta-gradient}
       \begin{split}
	\begin{aligned}
        \nabla_{\newgraph}^{meta} &=\nabla_{\newgraph}\mathcal{L}_{gen}\left(\hat{f}(\newgraph;\theta^*), {Y_P}\right) \\
     \text { s.t. } \quad \theta^* &= \underset\theta{\arg \min }\   \mathcal{L}_{train}\left(\hat{f}(\newgraph; \theta), Y_L\right). 
		\end{aligned}
	\end{split}
\end{align}


\begin{figure}[t!]
    \centering
\includegraphics[width=0.8\linewidth]{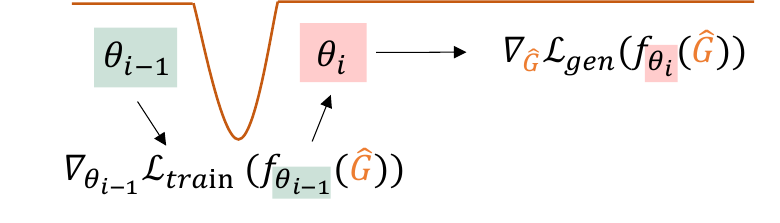}
\vspace{-0.1in}
    \caption{Computation of  the meta-gradient $\nabla_{\hat{G}}^{\mathrm{meta}}$. }
    \vspace{-0.1in}
    \label{fig:backpropagate}
\end{figure}

The meta-gradient indicates how the graph generation loss $\mathcal{L}_{gen}$ will change after training on the new graph. 
For simplicity, we use $\mathcal{L}_{gen}\newmodel_{\theta^*}(\newgraph) $ to denote $\mathcal{L}_{gen}(\hat{f}(\newgraph;\theta^*), {Y_P})$, 
and $\mathcal{L}_{train}\newmodel_{\theta}(\newgraph)$ to denote $\mathcal{L}_{train}(\hat{f}(\newgraph; \theta), Y_L)$, in the following discussions. 

\nop{
Then by treating the outer-level problem as the main optimization problem, the hyperparameter $\newgraph$ can be updated by leveraging the chain rule of differentiation:
\begin{equation}
    \label{eqn:chain}
    \nabla_{\newgraph}^{meta} = \nabla_{\newgraph}\mathcal{L}_{gen}\newmodel(\newgraph) + \nabla_{\theta}\mathcal{L}_{gen}\newmodel(\newgraph)\cdot\nabla_{\newgraph}\hat{\theta}.
\end{equation}
}

The optimization of the inner-level objective in Eqn. \eqref{equ:Bi-levl} can be achieved by multiple steps of
gradient descent starting from the initial parameter $\theta_0$ with the learning rate $\eta$: 
\begin{align} \label{equ:optimize the model parameters}
       \begin{split}
\begin{aligned}
\theta_{i+1}&=\theta_{i}-\eta\nabla_{\theta_{i}}\mathcal{L}_{train}\newmodel_{\theta_i}(\newgraph).
\end{aligned}
	\end{split}
\end{align}



Then the meta-gradient at the $i$-th iteration can be approximated as the follows (illustrated in Figure~\ref{fig:backpropagate}): 

\begin{align}
 \label{eqn:meta-prop}
       \begin{split}
	\begin{aligned}
\nabla_{\hat{G}}^{\mathrm{meta}} =& \frac{\partial \mathcal{L}_{\text {gen }}\left(\hat{f}_{\theta_{i}}(\hat{G})\right)}{\partial \hat{f}}\left[\frac{\partial \hat{f}_{\theta_{i}}(\hat{G})}{\partial \hat{G}}+\frac{\partial \hat{f}_{\theta_{i}}(\hat{G})}{\partial \theta_{i}} \frac{\partial \theta_{i}}{\partial \hat{G}}\right] \\=& \nabla_{\hat{f}} \mathcal{L}_{\operatorname{gen}}\left(\hat{f}_{\theta_{i}}(\hat{G})\right)\left[\nabla_{\hat{G}} \hat{f}_{\theta_{i}}(\hat{G})+\nabla_{\theta_{i}} \hat{f}_{\theta_{i}}(\hat{G}) \nabla_{\hat{G}} \theta_{i}\right]
		\end{aligned}
	\end{split}
\end{align}

where
\begin{align}
\label{eqn:chain-meta-gradient}
       \begin{split}
\begin{aligned}
\nabla_{\newgraph}\theta_{i}&=\nabla_{\newgraph}\theta_{i-1} -\eta\nabla_{\newgraph}\nabla_{\theta_{i-1}}\mathcal{L}_{train}(\hat{f}_{\theta_{i-1}}(\newgraph)). 
		\end{aligned}
	\end{split}
\end{align}


\subsection{Graph Reconstruction through Meta-gradient}

\begin{figure*}[h]
	\begin{center}
		\includegraphics[width=0.85\linewidth]{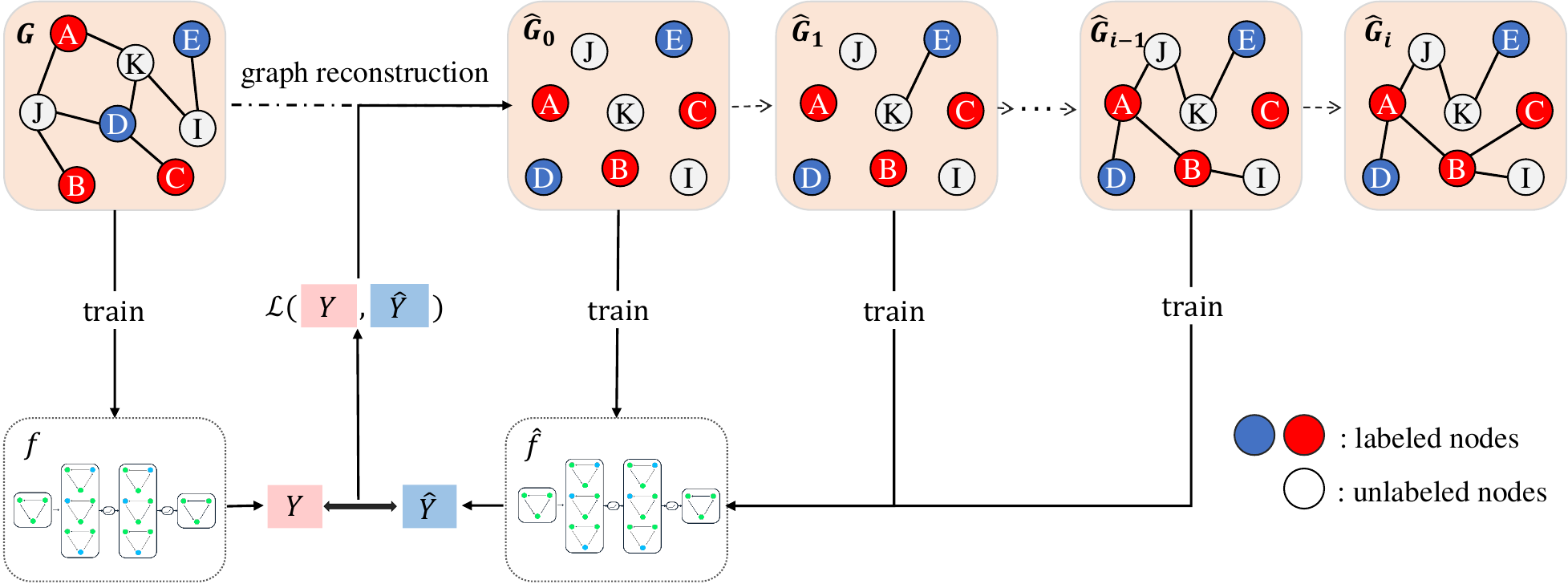}
        \caption{Illustration of \system. 
        In both  $\oldgraph$ and $\newgraph$, nodes $A$ - $E$ are labeled, with colors representing their respective classes.}  
         \label{fig:outline}  
	\end{center}
    \vspace{-0.1in}
\end{figure*}
We construct the graph $\hat{G}$ in an incremental fashion. Specifically, we initialize $\hat{G}$ as the  graph that contains all nodes in $G$ but no edges (i.e., all nodes are  disconnected in $\hat{G}^0$). Then at each iteration of the optimization of the inner-level objective in Eqn. \eqref{equ:Bi-levl},
we update the adjacency matrix $\hat{A}_{i+1}$ of $\hat{G}$ as follows:

\begin{align} \label{equ:gradient descend in A}
	\begin{split}
		\begin{aligned}
   \hat{A}_{i+1}=\hat{A}_{i}-\eta \cdot \nabla_{\newgraph}^{\text{meta}} 
		\end{aligned}
	\end{split}
\end{align} 

One problem of Eqn. \eqref{equ:gradient descend in A}  is that $\hat{A}_{i+1}$ is not a binary matrix. To fix this problem, first, we compute the complementary adjacency matrix $\bar{A} = J - A$, where $J$ is the all-ones matrix, and $A$ is the adjacency matrix of $\oldgraph$. Therefore, $\bar{A}[j,k]=1 - A[j,k]$ for each entry $A[j,k]$. 
Next, we compute a {\em score matrix} $g=\nabla^{meta}_{\newgraph}\odot \bar{A}$, obtained by performing element-wise production between the meta-gradients and $\bar{A}$.  
This operation ensures that the node pairs that are connected in $G$ will have score 0 in $g$. 
Then we greedily pick the nodes $j,k$ such that $g[j,k]$ has the lowest non-zero score in $g$:
 \begin{equation}
 e^*=\underset{e(j,k)\not\in\hat{A}_{i}}{\arg\min} g[j,k]
 \end{equation}
 
 Then we insert the edge $e(j,k)$ into $\hat{G}$ by changing $\hat{A}_{i+1}[j,k]=1$. 
 We choose this node pair as the insertion of its corresponding edge will minimize the loss on node classification. 
This completes the process of inserting one edge.  
The algorithm terminates after $\hat{K}$ iterations, where  $\hat{K}$ is a user-specified parameter for the number of edges in $\newgraph$. It guarantees that the reconstructed graph $\newgraph$ consists of $\hat{K}$ edges which do not appear in $\oldgraph$. 

\subsection{Implementation and Optimization} \label{sec:opt}

\begin{algorithm}[t!]
\caption{PGR algorithm}\label{algorithm1}
\KwIn{$\oldgraph$: Input  graph; $A$: the adjacency matrix of $\oldgraph$; $\oldmodel$: GNN model trained on $\oldgraph$; 
$\hat{K}$: number of edges in the constructed graph  $\newgraph$.}
$Y_L \leftarrow$ Labels of the labeled nodes in $\oldgraph$\; 
$Y_P \leftarrow$ Labels of unlabeled nodes in $\oldgraph$ predicted by $\oldmodel$\; 
$\bar{A}=J-A$\;
$\hat{A}_0 \leftarrow$ A self-connected diagonal matrix\;
$\theta_{0} \leftarrow$ Obtained by training on $\hat{A}_0$\;
\For{$i \in 0 \ldots \hat{K}-1$}
{
    
 $\theta_{i+1}=\theta_{i}-\eta \nabla_{\theta_{i}}  \mathcal{L}_{train}(\hat{f}_{{\theta_{i}}}(\hat{G}), Y_{L})$\; \label{alg:line8}
    
    $\g_{i} \leftarrow (\nabla_{\hat{G}} \mathcal{L}_{gen}(\hat{f}_{\theta_{i+1}}(\hat{G}), Y_P) \odot \bar{A}$\; \label{alg:line9}
    ${(j,k)} \leftarrow$ The index of the smallest gradient in $\g_{i}$\;
    $\hat{A}_{i}[j,k]=1$\;
}
\textbf{return} $\hat{G}, \theta_{\hat{K}}$.
\end{algorithm}

Putting everything together, Algorithm~\ref{algorithm1} sketches the flow of \system. 
At a high level, \system\ is implemented as an iterative process (as illustrated in Figure~\ref{fig:outline}). Specifically,  
\system\ trains a GNN $\oldmodel$ on $\oldgraph$, and obtains the labels $Y_P$ predicted by $\oldmodel$ (Line 2). Next, it 
initializes $\newgraph_0$ as a graph that contains nodes only (Line 4). 
At the $i$-th iteration, 
\system\ reconstructs a new graph $\newgraph_{i}$ using the meta-gradients (Lines 7 - 10) to insert one single edge. 
Therefore, \system\ inserts $\hat{K}$ edges by repeating $\hat{K}$ iterations. 

\smallskip
\noindent{\bf Efficient computation of meta-gradient.} We compute the meta-gradient using the built-in differentiation engine provided by  PyTorch~\cite{paszke2019pytorch} for computation of differentiation and gradients.  

\smallskip
\noindent{\bf One-step truncated back-propagation.} A commonly used approximate solution of meta-learning is {\em k-step truncated} {\em back-propagation} \cite{shaban2019truncated}, 
by which the inner model parameters $\theta$ is updated for $k$ steps rather than running to full convergence. In our implementation, we consider 1-step truncated back-propagation, i.e., Algorithm~\ref{algorithm1} performs a single iteration to update  $\hat{\theta}_{i+1}$ (Line 7). Our empirical results (Section  \ref{sc:factor}) will demonstrate that one-step truncated back-propagation is sufficient to effectively guide the optimization of the outer objective function of \system.  

\subsection{DP-PGR: Equip PGR with Edge-level DP}
\label{sc:dp-pgr}
Although \system\ significantly mitigates the topology privacy risks of GNNs against TIAs (see Section \ref{sec:evalution}), it has not provided formal privacy guarantee, even at the edge level. Therefore, in this section, we discuss how to equip \system\ with the edge-level DP guarantee. 
Note that our goal is not to design new DP algorithms. Instead, we aim to show that \system\ can be seamlessly integrated with existing edge-level DP methods and offer privacy protection at both edge and topology levels. 

\nop{
\begin{algorithm}
\caption{DP-PGR\Jeff{updated}\WW{The pseudo code does not help with presentation. Better remove. }}\label{alg:DP-PGR}
\KwIn{edge-DP mechanism $\mathcal{M(\cdot)}$, PGR algorithm $\mathcal{F(\cdot)}$, training dataset labels $Y_L$, feature matrix $X$, adjacency matrix $A$, privacy budget $\epsilon$.}
$P^{DP} \leftarrow \mathcal{M}(A, X, Y_L, \epsilon)$  \\
 $\hat{G}, \hat{\theta} \leftarrow \mathcal{F}(X, Y_L, P^{DP})$ \\
\textbf{return} $\hat{G}, \hat{\theta}$
\end{algorithm}
}

First, we formalize the definition of edge-level DP. 

\begin{definition}\label{def-Differential Privacy}
    ({\bf Edge-level Differential privacy~\cite{karwa2011private}}). A randomized algorithm $\mathcal{A}$ 
    satisfies $(\epsilon,\delta)$-edge differential privacy, if for any two adjacent graphs $G$ and $G'$ that differs in a single edge, and for any possible set of outputs $S\subseteq Range(\mathcal{A})$, we have: 
    \begin{align*}
        \forall 
        \mathrm{Pr}[\mathcal{A}(G)\in S]\leq e^{\epsilon}\mathrm{Pr}[\mathcal{A}(G^{\prime})\in S]+\delta.
    \end{align*}
\end{definition}

Here, $\epsilon$ is a non-negative parameter that quantifies the privacy loss, with smaller values indicating stronger privacy guarantees. The parameter $\delta \ge 0$ is a small probability that allows for a relaxation of the strict 
$\epsilon$-differential privacy condition. 
When $\delta=0$, $(\epsilon,\delta)$-DP becomes $\epsilon$-DP.

In general, existing edge-DP approaches can be broadly divided into two categories: 
1) {\em Graph synthesis (GS)} methods~\cite{tang2024edge,yuan2023privgraph,wu2022linkteller}  : these methods generate a synthesized graph that satisfies edge-DP. Then the GNN model is trained on the synthesized graph. 
2) {\em Noisy aggregation (NAG)} methods~\cite{kolluri2022lpgnet,sajadmanesh2023gap}: these methods add noise to feature aggregation during model training.

Due to the fundamental difference between these two categories, we adapt the two categories of edge-DP approaches to \system\ in different ways:  

{\bf GS-based edge-DP methods}: To equip edge-DP with \system, first, we replace $Y_P$ of Algorithm~\ref{algorithm1} (Line 2)
 with $Y_P^{DP}$, i.e., the labels output by the differentially private GNN models. Second, we replace $A$ in Algorithm~\ref{algorithm1} (Line 3) with $A^{DP}$, where $A^{DP}$ is the adjacency matrix of the  graph generated by the GS-based methods.  

{\bf NAG-based edge-DP methods}: Similar to the GS-based methods, $Y_P$ of Algorithm~\ref{algorithm1} (Line 2)
 will be replaced with $Y_P^{DP}$. However, as these methods do not generate a graph that satisfies edge-DP, we cannot replace $A$ in Algorithm~\ref{algorithm1} (Line 3) with a graph that satisfies edge-DP. Consequently, we cannot compute $g_i$ from the complementary graph (Line 8). Instead, we pick the smallest gradient from $\g_{i} = (\nabla_{\hat{G}} \mathcal{L}_{gen}(\hat{f}_{\theta_{i+1}}(\hat{G}), Y_P)$. Doing this cannot guarantee that $\hat{G}$ is disjoint from $G$, thus bringing potential downgrade in topology privacy protection. However, this is the price we have to pay for achieving edge-level DP. 

\nop{
\begin{ditemize}
    \item  {\bf Adapt edge-DP to model training of Algorithm \ref{algorithm1}.}  As both types of edge-DP algorithms can yield a model whose output $P^{DP}$ is protected with the edge-DP guarantee, we replace $Y_P$ of Algorithm~\ref{algorithm1} (Line 2)
 with $Y^{DP}$. 

 \item {\bf Adapt edge-DP to graph overlap evaluation of Algorithm \ref{algorithm1}.}
 Regarding GS-based edge-DP methods, In terms of the NAG-based edge-DP methods, we cannot directly apply these methods to \system\ as they do not generate such a graph satisfy edge-DP. Therefore, after obtaining the meta-gradient of $\hat{G}$, we do not  perform the element-wise product with $\bar{A}$ (Line 8), so that there is no access to the private graph. \WW{Discuss that no access to the complementary graph cannot guarantee non-overlap requirement, sacrificing on topology privacy protection. }
\end{ditemize}
}

\smallskip
We have the following theorem. 

\begin{theorem}
\label{theorem:DP-PGR}
DP-PGR satisfies $(\epsilon,\delta)$ edge-DP.
\end{theorem}

The proof of Theorem \ref{theorem:DP-PGR} can be found in Appendix \ref{appendix:proofdp}. 
To summarize, the design of DP-PGR shows that \system\ can be easily integrated into the existing edge-DP methods, providing privacy protection at both the edge and topology levels. 

\nop{
\begin{align}
       \begin{split}
	\begin{aligned}
\nabla_{\hat{G}}{\theta}_{i+1}&=\nabla_{\hat{G}}{\theta}_{i}-\eta\nabla_{\hat{G}}\nabla_{{\theta}_{i}}\mathcal{L}_{train}(f_{\theta_i}(\newgraph)) \\
&=-\eta\nabla_{\hat{G}}\nabla_{{\theta}_{i}}\mathcal{L}_{train}(f_{\theta_i}(\newgraph)).
		\end{aligned}
	\end{split}
\end{align}
}

\subsection{Robustness of \system\ against Grey-box Adversary}
\label{sc:grey-box}

\begin{figure}[t!]
	\begin{center}
\includegraphics[width=0.8\linewidth]{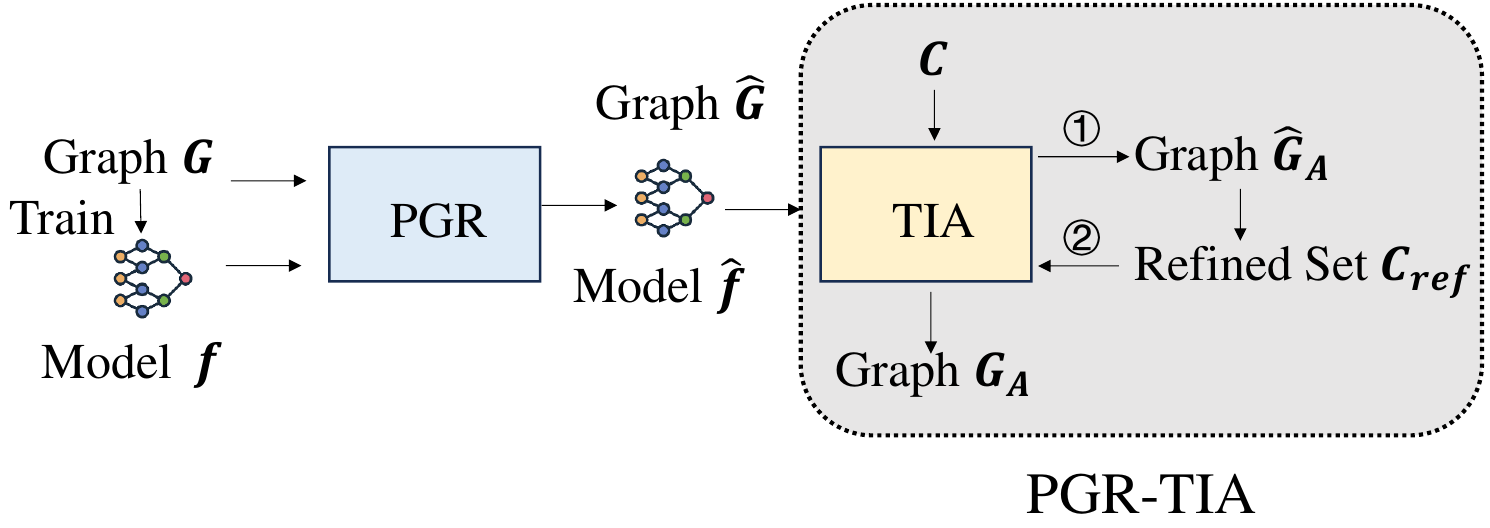}
		\caption{Illustration of PGR-TIA.} \vspace{-0.1in}
        \label{figure:topology privacy loss}
	\end{center}
    \vspace{-0.1in}
\end{figure}

In this section, we discuss the robustness of \system\ in the presence of adversaries who try to bypass it. We consider a {\em grey-box adversary} who has not only black-box access to the trained GNN model $\newmodel$ but also some partial knowledge of \system, including its mechanisms and the value of its parameters such as $\hat{K}$. 
The adversary will exploit the knowledge of \system\ to enhance the TIA accordingly. Following this intuition, we design {\bf PGR-TIA}, an enhanced TIA against \system. Next, we present the details of PGR-TIA. 

Notably, \system\ possesses a key property: {\em the training graph $\newgraph$ used for the output model $\newmodel$ shares no edges with the original training graph $\oldgraph$ used to train $\oldmodel$}. This property enables a grey-box adversary to design stronger TIAs. PGR-TIA is tailored to exploit this edge-disjointness information. Importantly, PGR-TIA does not require access to the original graph $\oldgraph$ or model $\oldmodel$ - it operates solely via black-box access to $\newmodel$.

Specifically, let $\oldgraph$ denote the original graph and $\oldmodel$ the GNN trained on it. Let $\newgraph$ and $\newmodel$ represent the graph and model generated by \system, respectively. Define the set of all possible node pairs (i.e., candidate edges) in $\oldgraph$ as $C = \{(u, v) \mid (u, v) \in G\}$. Essentially, PGR-TIA consists of two rounds of TIAs. In the first round, the adversary performs a black-box attack on $\newmodel$ to infer a graph $\hat{G}_{A}$, with $\hat{E}_A$ denoting its edge set.  
Since the adversary knows that it is highly likely that $\hat{E}_{A} \cap E = \emptyset$, they can refine the candidate edge set $C$ by excluding the edges inferred in $\hat{G}_{A}$. That is, the refined candidate set is defined as $C_{\text{ref}} = C \setminus \hat{C}_{A}$, where $\hat{C}_{A}$ includes all node pairs connected in $\hat{G}_{A}$. Using this refined set, the adversary then conducts the second round of TIA, selecting edges exclusively from $C_{\text{ref}}$ to infer $G_{A}$.

Compared to traditional TIAs, \textsf{PGR-TIA} reduces the edge search space from $\frac{N(N-1)}{2}$ to $\left(\frac{N(N-1)}{2} - \hat{K}\right)$, where $\hat{K}$ denotes the number of edges in $\newgraph$ - a tunable parameter in \system. Intuitively, a larger $\hat{K}$ offers stronger pruning power, leading to a more efficient attack. However, the effectiveness of this refinement depends heavily on the accuracy of the TIA. The pruned edges are guaranteed to be correct only if $\hat{G}_{A} = \newgraph$, i.e., the TIA achieves perfect accuracy—an unrealistic assumption in practice. When the TIA is imperfect, some edges in $\hat{E}_{A}$ may still exist in $E$, resulting in true edges being incorrectly excluded from the refined candidate set. These exclusions can degrade the final accuracy of the inferred graph $G_{A}$. Intuitively, a smaller $\hat{K}$ leads to a higher likelihood of such erroneous exclusions. For this reason, we prefer a smaller $\hat{K}$, as it enhances \system's robustness against \textsf{PGR-TIA}.

\vspace{-0.1in}
\section{Evaluation} \label{sec:evalution}
In this section, we evaluate the performance of \system\  by answering the following five research questions:
\begin{ditemize}
    \item $\mathbf{RQ_1}$ - How effective is \system\ against TIAs?
    \item $\mathbf{RQ_2}$ - How does \system\ perform compared to the existing edge-DP approaches in terms of the trade-off between topology privacy protection and model accuracy?
    \item $\mathbf{RQ_3}$ - How do various parameters impact \system's performance?
    \item $\mathbf{RQ_4}$ - How robust is \system\ against the PGR-TIA attack?
    \item $\mathbf{RQ_5}$ - How does DP-PGR perform compared to existing edge-DP approaches in terms of both topology privacy protection and model accuracy?
\end{ditemize}

\subsection{Experiment Setup} \label{sub:experiment setup}
\vspace{-0.05in}

All the experiments are executed on a server with NVIDIA A100
GPU and 40GB memory. All the algorithms are implemented in Python along with PyTorch. 
All results were evaluated as the average of three trials. 

\smallskip
{\bf Datasets.} Our experiments utilize four real-world graph datasets, including two citation graphs (Cora~\cite{yang2016revisiting} and Citeseer~\cite{yang2016revisiting}) and two social network graphs (LastFM~\cite{rozemberczki2020characteristic} and Emory~\cite{traud2012social}). The statistics of these datasets are shown in 
Table~\ref{table:Statistics of datasets}.
For each dataset, 10\% of nodes are randomly selected as the labeled nodes for training, and the remaining 90\%  nodes are used for testing. 

\smallskip
\begin{table}[t!]
\small
	\centering
	\caption{Statistics of datasets}
\begin{tabular}{llccccc}
\toprule
\textbf{Type}                                                                & \textbf{Dataset} & \textbf{Nodes} & \textbf{Edges} & \textbf{Density} & \textbf{Attrs} & \textbf{Classes} \\ \hline
\multirow{2}{*}{\begin{tabular}[c]{@{}c@{}}Citation\\  Network\end{tabular}} & Cora            & 2708           & 5429           & 0.14\%           & 1433           & 7                \\
                                                                             & Citeseer         & 3327           & 4732           & 0.08\%           & 3703           & 6                \\ \hline
\multirow{2}{*}{\begin{tabular}[c]{@{}c@{}}Social \\ Network\end{tabular}}   & LastFM           & 7624           & 55612          & 0.19\%           & 128            & 18               \\
                                                                             & Emory            & 5125           & 425138         & 3.23\%           & 288            & 5               
                                                                                   \\ \bottomrule
\end{tabular}
	\label{table:Statistics of datasets}
\end{table}

{\bf GNN models.} We consider three mainstream GNN models, namely, Graph Convolutional Networks (GCN)~\cite{traud2012social}, Graph Attention Networks (GAT)~\cite{petar2018graph}, and GraphSAGE~\cite{hamilton2017inductive}. We set up both GCN and GraphSAGE as two-layer neural networks with a 32-neuron hidden layer, and GAT as  a 2-layer network with an 8-neuron hidden layer. We utilize the ReLU activation function and Softmax function for the output. We employ the Adam optimizer for training, with a learning rate of $0.01$, and a weight decay of $5e-4$. All the three models were trained by $100$ epochs.\footnote{Our empirical results show that 100 epochs are sufficient, as the model can achieve satisfactory accuracy. Specifically, with GCN model, the model's accuracy on the Cora, Citeseer, LastFM, Emory, and Facebook datasets is 83.38\%, 70.31\%, 81.75\%, 90.68\%, and 91.08\%, respectively. For GTA, we train 200 epochs on Emory.}

\begin{table*}[t!]
\centering
\small
\caption{Topology privacy leakage (TPL) and model accuracy loss of \system. We consider three TIAs for measuring TPL. A positive (negative, resp.) loss indicates that the node classification accuracy is downgraded (upgraded, resp.) by privacy protection.}
\begin{tabular}{c|c|cccccc|c}
\hline
\multirow{3}{*}{\textbf{Model}} & \multirow{3}{*}{\textbf{Dataset}} & \multicolumn{6}{c|}{\textbf{Topology privacy leakage (\%)}}                                                                                                                                                  & \multirow{3}{*}{\textbf{Model accuracy loss (\%)}} \\ \cline{3-8}
                                &                                   & \multicolumn{3}{c|}{\textbf{Original model}}                                                                    & \multicolumn{3}{c|}{\textbf{PGR}}                                                          &                                                    \\ \cline{3-8}
                                &                                   & \multicolumn{1}{c|}{\textbf{M-TIA}} & \multicolumn{1}{c|}{\textbf{C-TIA}} & \multicolumn{1}{c|}{\textbf{I-TIA}} & \multicolumn{1}{c|}{\textbf{M-TIA}} & \multicolumn{1}{c|}{\textbf{C-TIA}} & \textbf{I-TIA} &                                                    \\ \hline
\multirow{3}{*}{GCN}            & Cora                              & \multicolumn{1}{c|}{28.10}           & \multicolumn{1}{c|}{25.53}          & \multicolumn{1}{c|}{100.00}            & \multicolumn{1}{c|}{3.01}           & \multicolumn{1}{c|}{2.70}            & 0.81           & 0.00                                                  \\ \cline{2-9} 
                                & Citeseer                          & \multicolumn{1}{c|}{17.33}          & \multicolumn{1}{c|}{10.73}          & \multicolumn{1}{c|}{100.00}            & \multicolumn{1}{c|}{4.17}           & \multicolumn{1}{c|}{2.30}            & 1.14           & 0.00                                                  \\ \cline{2-9} 
                                & Emory                             & \multicolumn{1}{c|}{17.55}          & \multicolumn{1}{c|}{16.54}          & \multicolumn{1}{c|}{100.00}            & \multicolumn{1}{c|}{12.53}          & \multicolumn{1}{c|}{12.01}          & 9.32           & 0.38                                               \\ \hline
\multirow{3}{*}{GAT}            & Cora                              & \multicolumn{1}{c|}{14.85}          & \multicolumn{1}{c|}{11.50}           & \multicolumn{1}{c|}{100.00}            & \multicolumn{1}{c|}{2.71}           & \multicolumn{1}{c|}{2.44}           & 1.08           & -0.12                                              \\ \cline{2-9} 
                                & Citeseer                          & \multicolumn{1}{c|}{20.45}          & \multicolumn{1}{c|}{11.66}          & \multicolumn{1}{c|}{100.00}            & \multicolumn{1}{c|}{4.81}           & \multicolumn{1}{c|}{4.22}           & 3.51           & 0.00                                                  \\ \cline{2-9} 
                                & Emory                             & \multicolumn{1}{c|}{16.76}          & \multicolumn{1}{c|}{17.44}          & \multicolumn{1}{c|}{100.00}            & \multicolumn{1}{c|}{12.03}          & \multicolumn{1}{c|}{11.34}          & 10.44          & 0.43                                               \\ \hline
\multirow{3}{*}{GraphSAGE}      & Cora                              & \multicolumn{1}{c|}{13.13}          & \multicolumn{1}{c|}{9.68}           & \multicolumn{1}{c|}{83.15}          & \multicolumn{1}{c|}{4.10}            & \multicolumn{1}{c|}{2.40}            & 4.74           & 0.13                                               \\ \cline{2-9} 
                                & Citeseer                          & \multicolumn{1}{c|}{17.57}          & \multicolumn{1}{c|}{17.75}          & \multicolumn{1}{c|}{100.00}            & \multicolumn{1}{c|}{4.50}            & \multicolumn{1}{c|}{4.20}            & 3.51           & 0.00                                                  \\ \cline{2-9} 
                                & Emory                             & \multicolumn{1}{c|}{18.79}          & \multicolumn{1}{c|}{17.62}          & \multicolumn{1}{c|}{94.86}          & \multicolumn{1}{c|}{13.21}          & \multicolumn{1}{c|}{12.45}          & 10.10           & 0.87                                               \\ \hline
\end{tabular}
\label{table:performance}
\end{table*}


\smallskip

{\bf Evaluation metrics.}
We consider following evaluation metrics:
\begin{ditemize}
\item \emph{Topology privacy vulnerability:}
We utilize TPL (Eqn. \eqref{equ:first part loss}) to evaluate the effectiveness of the defense mechanism.  A higher (lower, resp.) TPL value indicates weaker (stronger, resp.) defense.  
\item \emph{Model accuracy loss:} We measure the {\em loss of model accuracy} as $\frac{ACC'- ACC}{ACC}$,
where $ACC$ and $ACC'$ are the accuracy of node classification  before and after the deployment of the defense. The model accuracy loss can be either positive or negative: a positive (negative, resp.) loss indicates that the node classification accuracy is downgraded (upgraded, resp.) by the privacy protection mechanism. 
\end{ditemize}

\smallskip

{\bf Methods for comparison.} We compare \system\  with six existing graph privacy protection approaches: Eclipse~\cite{tang2024edge}, LPGNet~\cite{kolluri2022lpgnet}, PrivGraph~\cite{yuan2023privgraph}, GAP~\cite{sajadmanesh2023gap}, LapEdge~\cite{wu2022linkteller} and EdgeRand~\cite{wu2022linkteller}.\footnote{We do not consider PPNE~\cite{han2023privacy} and PrivGNN~\cite{olatunjireleasing} as both PPNE and PrivGNN assume the availability of either a partial graph or an external dataset, which do not apply to our setting. 
}  
Among these methods, Eclipse, PrivGraph, LapEdge, and EdgeRand generate synthetic graphs for training the GNN model, while LPGNet and GAP perturb the GNN training process.  We use the privacy budget $\epsilon \in \{1,3,5,7,9\}$ for these approaches.

\begin{figure*}[h]
	\begin{center}
		\includegraphics[width=1.0\linewidth]{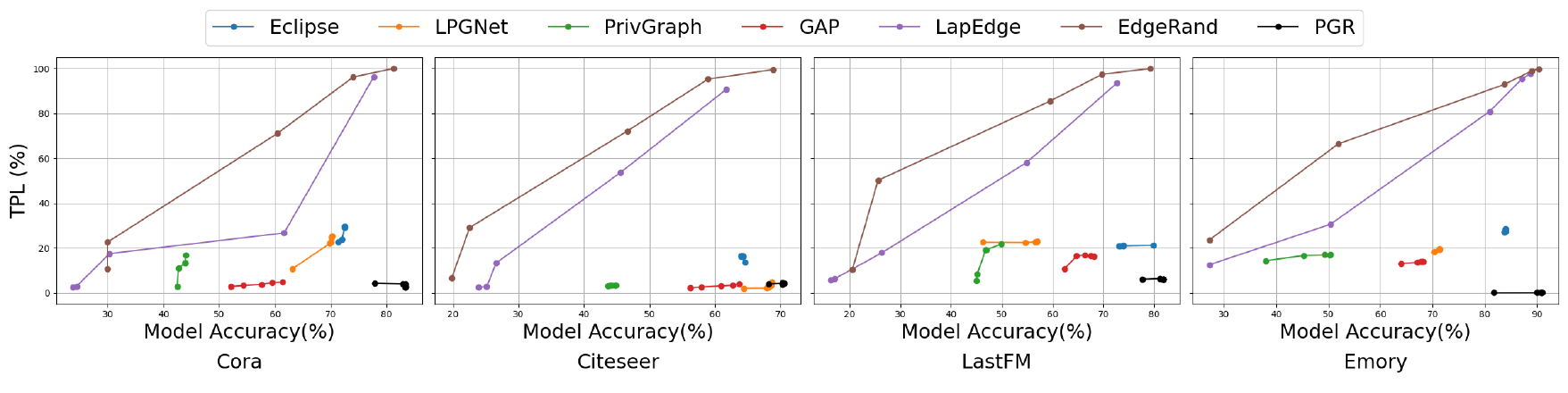}
        \vspace{-0.2in}
		\caption{The trade-off between topology privacy leakage and model accuracy. The x-axis represents node classification accuracy (\%), while the y-axis represents topology privacy leakage ($TPL$). For all edge-dp methods, we set the privacy budget $\epsilon$ in \{1,3,5,7,9\}.} 
        \label{figure:tradeoff_auc_acc}
	\end{center}
    \vspace{-0.1in}
\end{figure*}

\subsection{Effectiveness of \system\  ($\mathbf{RQ_1}$)}

Table~\ref{table:performance} presents the TPL results against all three TIAs, both before and after the deployment of \system. Overall, \system\ significantly reduces topology privacy leakage across all attacks, datasets, and models, demonstrating its effectiveness as a defense mechanism. Notably, \system\ is particularly effective against I-TIA - the strongest of the three attacks - reducing TPL from approximately 100\% to no more than 5\% for both Cora and Citeseer datasets, and around 10\% for the Emory dataset. 
For M-TIA and C-TIA, although the reductions in TPL are less dramatic compared to I-TIA, \system\ still achieves substantial mitigation. One exception is the Emory dataset, where \system\ does not lower the TPL to the same extent as in other datasets. However, this is acceptable, as the resulting TPL in this case is already close to the level of random guessing. The variation in \system's defense effectiveness across datasets can be attributed to differences in graph density, which we further analyze in Section~\ref{sc:factor}.


Additionally, while offering strong protection for topology privacy, \system\ incurs minimal loss in model accuracy, with the maximum accuracy degradation not exceeding 0.87\%. In rare cases, we even observe that \system\ slightly outperforms the original model by 0.12\% in accuracy.

In summary, \system\ provides topological privacy protection while preserving the model performance, addressing the insufficiency of existing edge-DP approaches.




\subsection{Trade-off between Privacy and Model Accuracy ($\mathbf{RQ_2})$}

Privacy often comes at the expense of reduced model utility. To illustrate the trade-off between privacy and model accuracy, we construct a TPL-utility curve, where each point represents a pair of topology privacy leakage ($TPL$) and node classification accuracy values. This curve is generated by varying the parameter $\hat{K}$ of \system\ as well as the privacy budget $\epsilon$ of the baselines, and measuring both the $TPL$  and node classification accuracy for each parameter setting. Intuitively, a method that achieves lower $TPL$ and higher node classification accuracy strikes a better balance between privacy and model performance.

Figure~\ref{figure:tradeoff_auc_acc} demonstrates the TPL-utility curve of \system\ and the six existing edge-DP approaches on four datasets. 
Intuitively, the defense method that has higher defense strength (i.e. lower TPL) and aligned target model accuracy with the others has a better trade-off between privacy and model accuracy.  
We observe that \system\ outperforms the six baselines in the trade-off between defense effectiveness and model accuracy:  \system\ achieves lower TPL than the baselines under the same target model accuracy, while it delivers higher target model accuracy than the baselines when they have the same attack accuracy.

\subsection{Factor Analysis ($\mathbf{RQ_3}$)}
\label{sc:factor}

In this part of the experiment, we investigate the impact of several key factors on the performance of \system, including: (1) the number of edges $\hat{K}$ in the reconstructed graph $\newgraph$, which is a tunable parameter in \system; (2) the structural characteristics of the original training graph $\oldgraph$; and (3) the number of iterations used for optimizing the inner objective function in Algorithm~\ref{algorithm1}. We report the highest TPL observed across the three TIAs, along with the corresponding model accuracy. Additionally, we evaluate the effect of varying the number of GNN layers on \system’s performance, with detailed results
 provided in Appendix~\ref{appendix:numberoflayers}.

\begin{figure*}[htb]
	\begin{center}
		\includegraphics[width=1.0\linewidth]{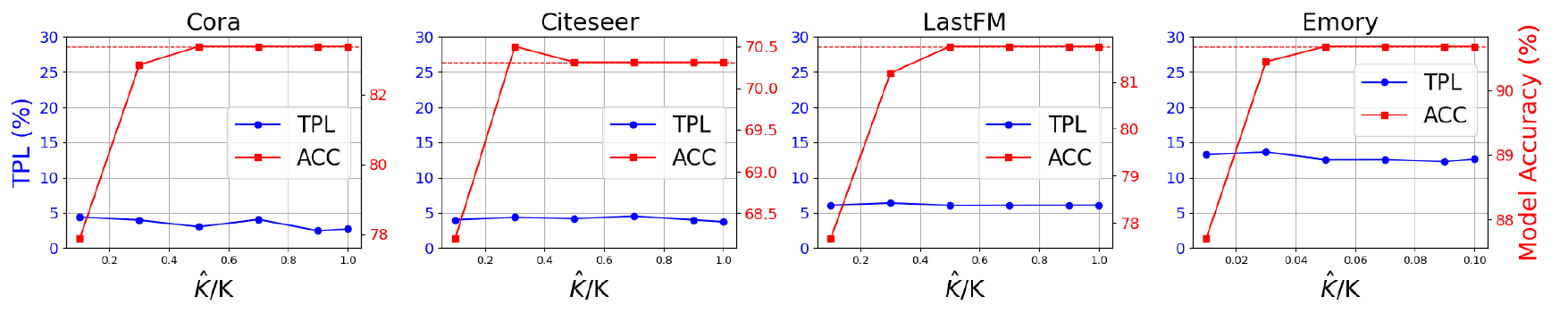}
        \vspace{-0.2in}
		\caption{Impact of the number of edges $\hat{K}$ on the performance of \system. The x-axis indicates the proportion $\hat{K}/K$ ($K$: number of real edges). The left (blue) y-axis indicates the topology privacy leakage ($TPL$), while the right (red) y-axis indicates model  accuracy. The dotted red line denotes the accuracy of the original model without privacy protection.}
        \label{figure:impact_k_hat}
	\end{center}
\end{figure*}

\smallskip
\noindent{\bf Number of edges $\hat{K}$.} 
Figure~\ref{figure:impact_k_hat} shows the results of both topology privacy leakage (TPL) and model accuracy under varying values of $\hat{K}$, the number of edges in the reconstructed graph. Instead of fixing $\hat{K}$, we express it as a proportion of $K$, the number of edges in the original training graph $\oldgraph$. 
The results reveal that TPL remains relatively stable as $\hat{K}$ increases, highlighting the robustness of \system's defense capability. In contrast, model accuracy improves with larger $\hat{K}$. Notably, \system\ does not require a large $\hat{K}$ to achieve strong performance. Even when $\newgraph$ contains significantly fewer edges than $\oldgraph$, it can still support effective node classification. For example, on the Citeseer dataset, a reconstructed graph with only 30\% of $\oldgraph$'s edges achieves comparable accuracy. This efficiency is even more pronounced on denser graphs---on the Emory dataset, just 3\% of $\oldgraph$'s edges are sufficient to maintain similar performance.


\begin{figure*}[t!]
    \centering
    \begin{subfigure}[b]{0.24\textwidth}
        \includegraphics[width=\textwidth]{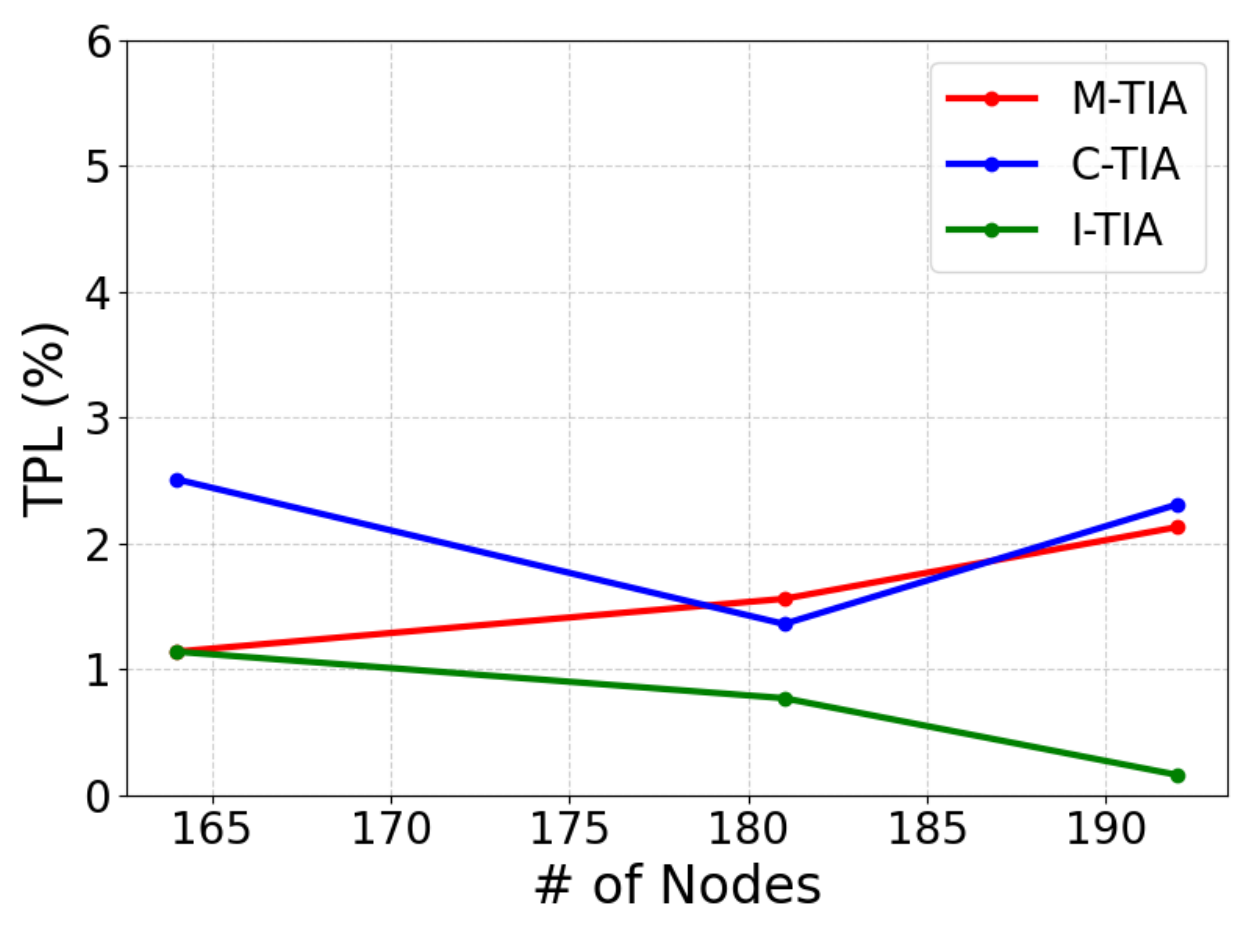} 
        \caption{Number of nodes}
        \label{fig:number_of_nodes}
    \end{subfigure}
    \begin{subfigure}[b]{0.24\textwidth} 
        \includegraphics[width=\textwidth]{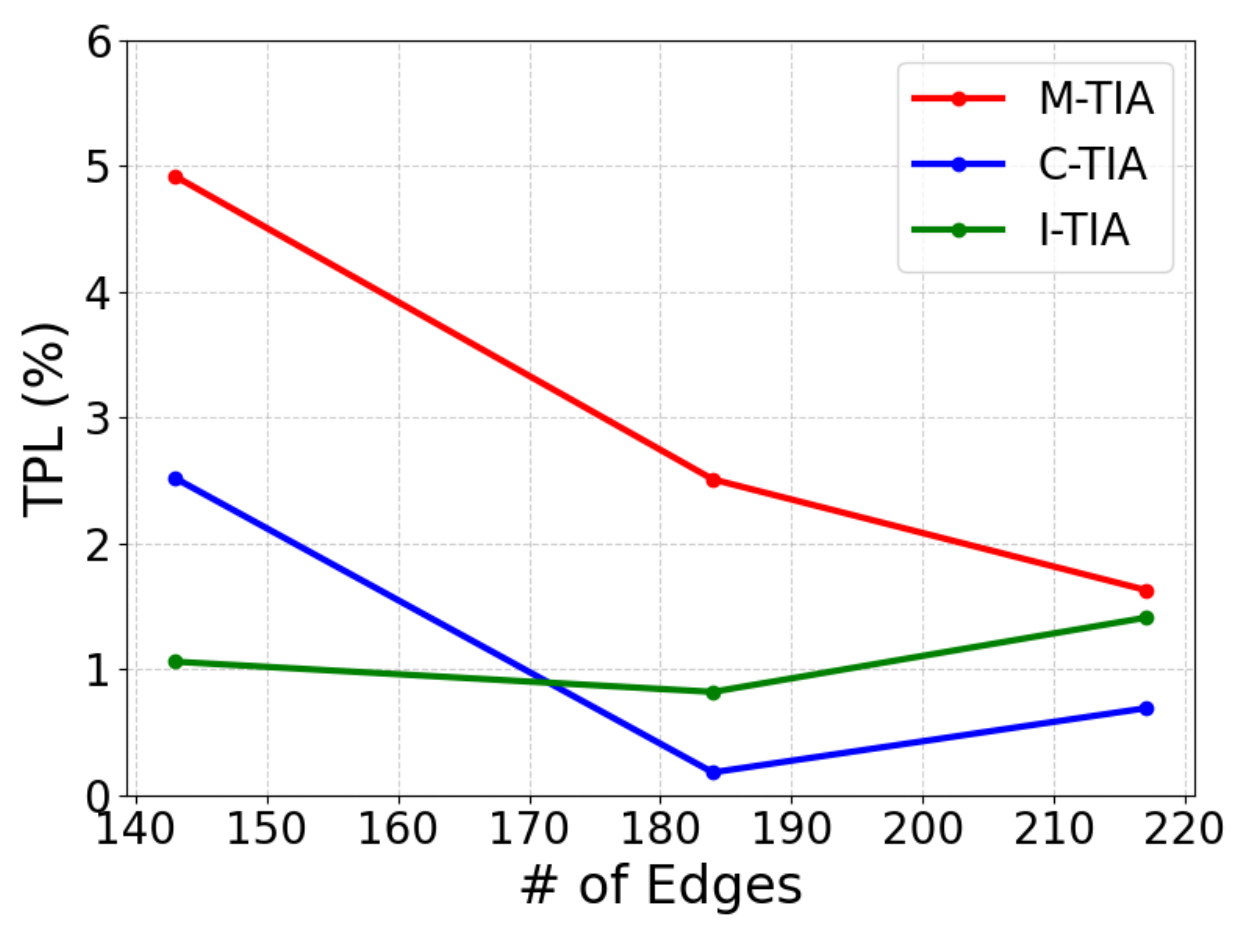} 
        \caption{Number of edges}
        \label{fig:number_of_edges}
    \end{subfigure}
    \begin{subfigure}[b]{0.24\textwidth}
        \includegraphics[width=\textwidth]{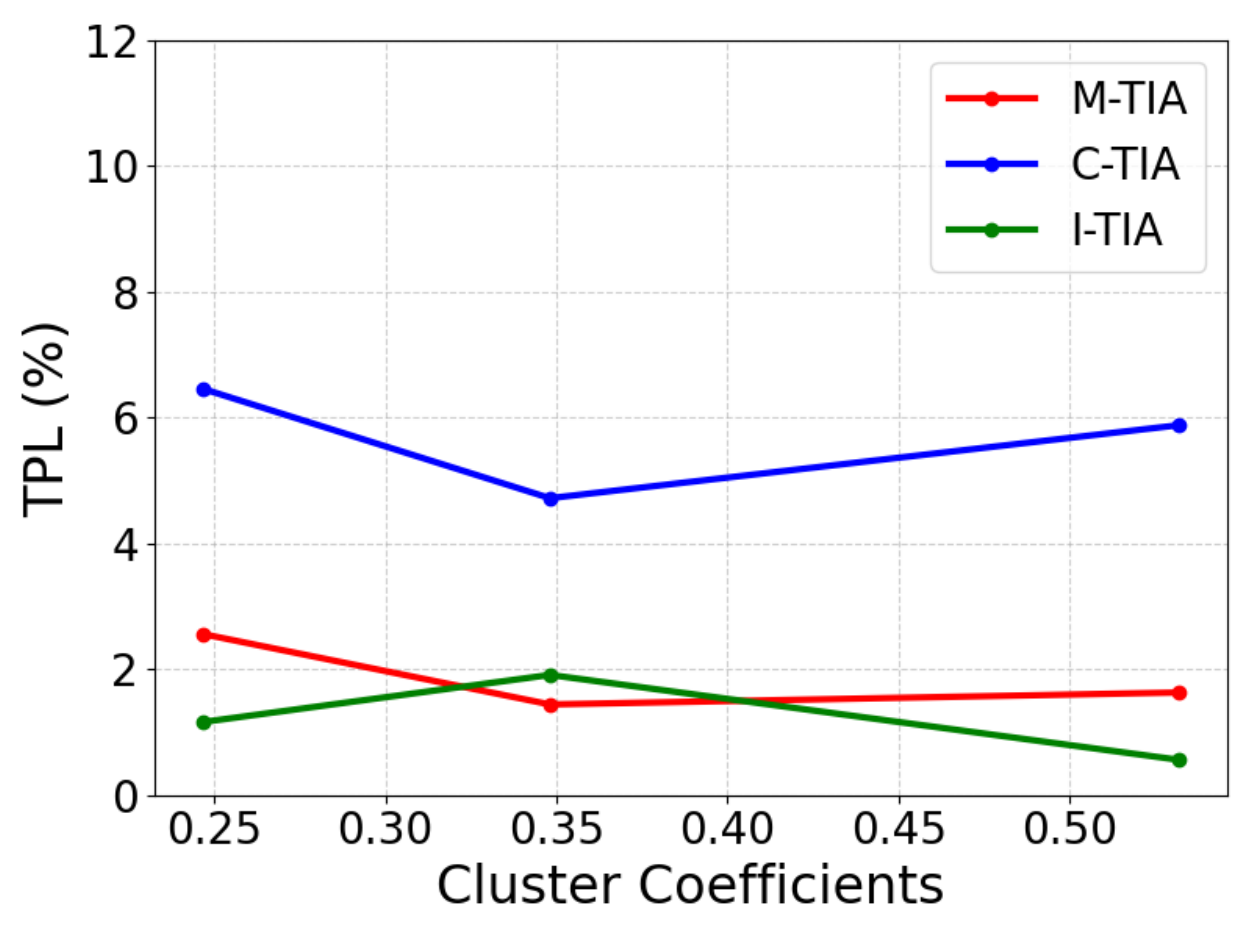}
    \caption{Cluster coefficient}
        \label{fig:number_of_cluster_coefficients}
    \end{subfigure}
     \begin{subfigure}[b]{0.24\textwidth}
        \includegraphics[width=\textwidth]{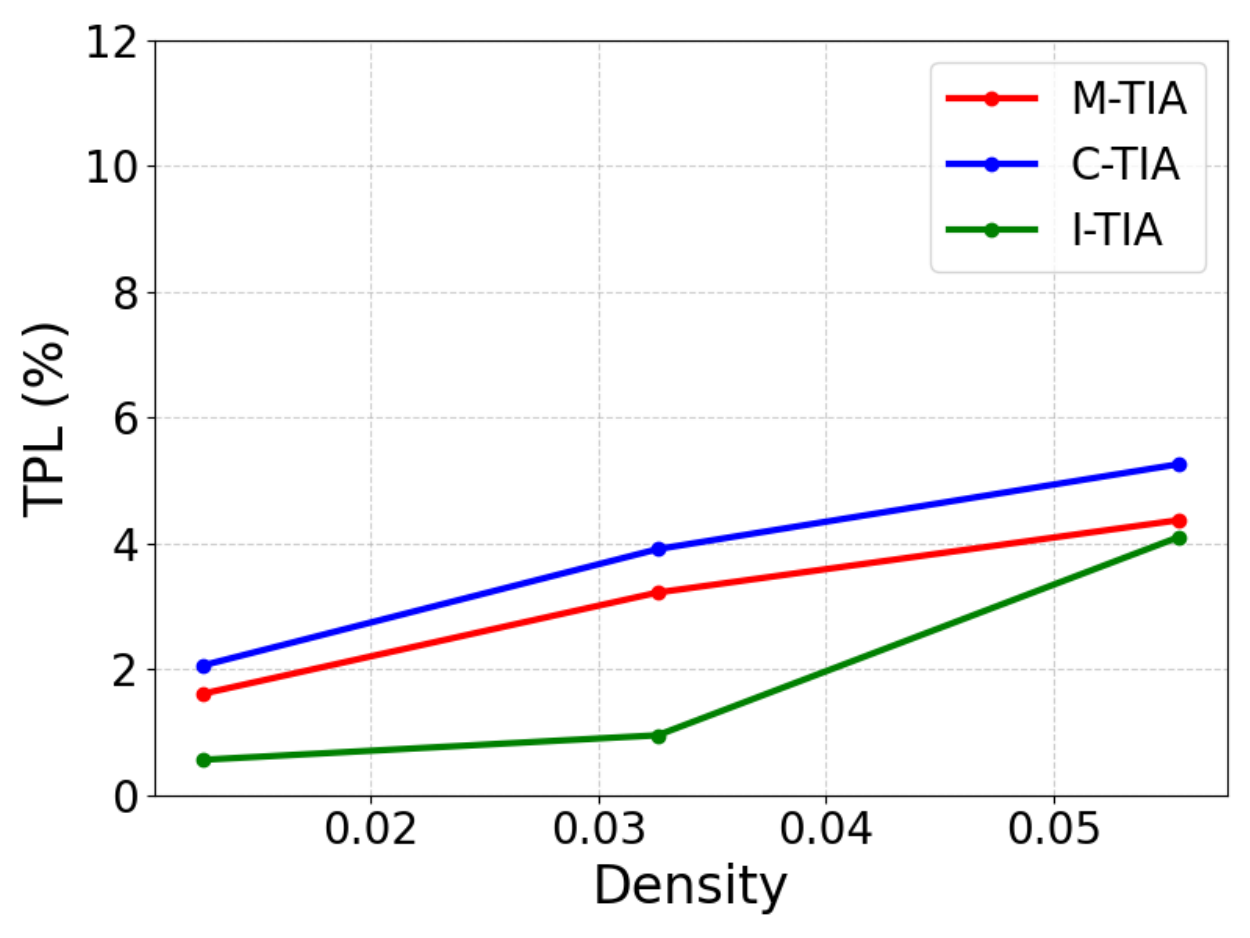}
        \caption{Graph density}
\label{fig:number_of_density}
    \end{subfigure}
    \vspace{-0.1in}
     \caption{The impact of graph characteristics on the performance of \system\ (Cora dataset).} 
    \label{figure:graph_characteristics}
\end{figure*}

\smallskip
\noindent{\bf Graph characteristics.}  In our previous results, we observed that \system's performance varies across different graphs. To better understand how specific graph characteristics influence its effectiveness, we investigate four key graph measures: {\em number of nodes}, {\em number of edges}, {\em density}, and {\em clustering coefficient}. To isolate the impact of each factor, we randomly sample subgraphs from the Cora dataset, keeping the number of nodes fixed while varying the other three properties.
Figure~\ref{figure:graph_characteristics} presents the TPL results of \system\ under these settings. The TPL does not exhibit consistent changes when the three types of graph measures  — edge count, node count, and clustering coefficient— are varied. This suggests that \system\ remains effective across datasets of varying sizes and community structures. In contrast, we 
observe a consistent pattern in TPL changes as graph density increases (Figure~\ref{figure:graph_characteristics} (d)). Specifically, \system\ performs better on sparser graphs, exhibiting lower TPL values. This aligns with our earlier observation that \system\ achieves the highest TPL on the Emory dataset, which is the densest among all the datasets.

\smallskip
\noindent{\bf One-step truncated back-propagation optimization.} 
Recall that we use the one-step truncated back-propagation to approximate the parameters (Section~\ref{sec:opt}). In this part of experiments, we evaluate the impact of this optimization strategy on \system's performance. We define convergence as the condition where the change in the loss remains within $0.01$ over five consecutive iterations.
Table~\ref{tab:num_of_inner_iteration} compares the performance of \system\ under two settings: (1) performing only a single update iteration for the inner optimization loop (Line 9 of Algorithm~\ref{algorithm1}), and (2) training the model until the inner objective converges. The results reveal that \system\ achieves comparable performance in both settings—model accuracy and TPL remain similar regardless of whether a single iteration or full convergence is used. This validates our design choice and highlights the efficiency of the proposed optimization approach in Algorithm~\ref{algorithm1}. 

\begin{table}[t!]
\small
\centering
\caption{Impact of one-step truncated back-propagation on \system's performance. The highest TPL among the three TIAs is reported.}
\vspace{-0.1in}
\begin{tabular}{c|cc|cc|cc|cc}
\hline
\multirow{2}{*}{\bf Setting}& \multicolumn{2}{c|}{\bf Cora}         & \multicolumn{2}{c|}{\bf Citeseer}     & \multicolumn{2}{c|}{\bf LastFM}       & \multicolumn{2}{c}{\bf Emory}          \\ \cline{2-9} 
& \multicolumn{1}{c|}{\bf ACC}   & {\bf TPL}  & \multicolumn{1}{c|}{\bf ACC}   & {\bf TPL}  & \multicolumn{1}{c|}{\bf ACC}   & {\bf TPL}  & \multicolumn{1}{c|}{\bf ACC}   & {\bf TPL}   \\ \hline
Conv.   & \multicolumn{1}{c|}{83.38} & 3.21 & \multicolumn{1}{c|}{70.31} & 3.62 & \multicolumn{1}{c|}{81.75} & 6.14 & \multicolumn{1}{c|}{91.07} & 12.57 \\ \hline
1 iter.            & \multicolumn{1}{c|}{83.38} & 3.01 & \multicolumn{1}{c|}{70.31} & 4.17 & \multicolumn{1}{c|}{81.75} & 6.07 & \multicolumn{1}{c|}{90.75} & 12.53 \\ \hline
\end{tabular}
\label{tab:num_of_inner_iteration}
\end{table}

\begin{figure}[t!]
    \centering
    \begin{subfigure}[b]{0.23\textwidth} 
        \includegraphics[width=\textwidth]{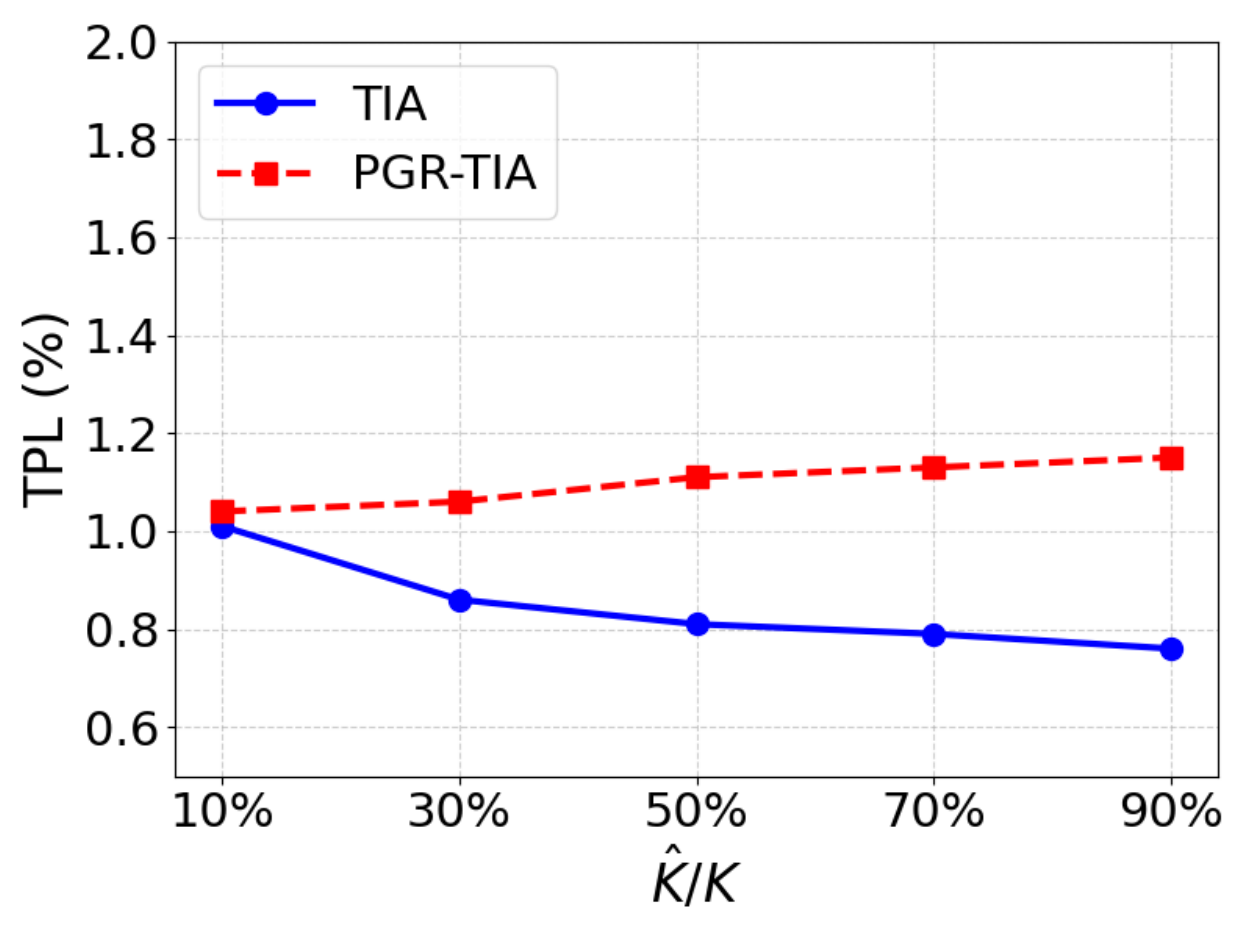} 
        \caption{Cora}
        \label{fig:PGR_TIA_cora}
    \end{subfigure}
    \begin{subfigure}[b]{0.23\textwidth}
        \includegraphics[width=\textwidth]{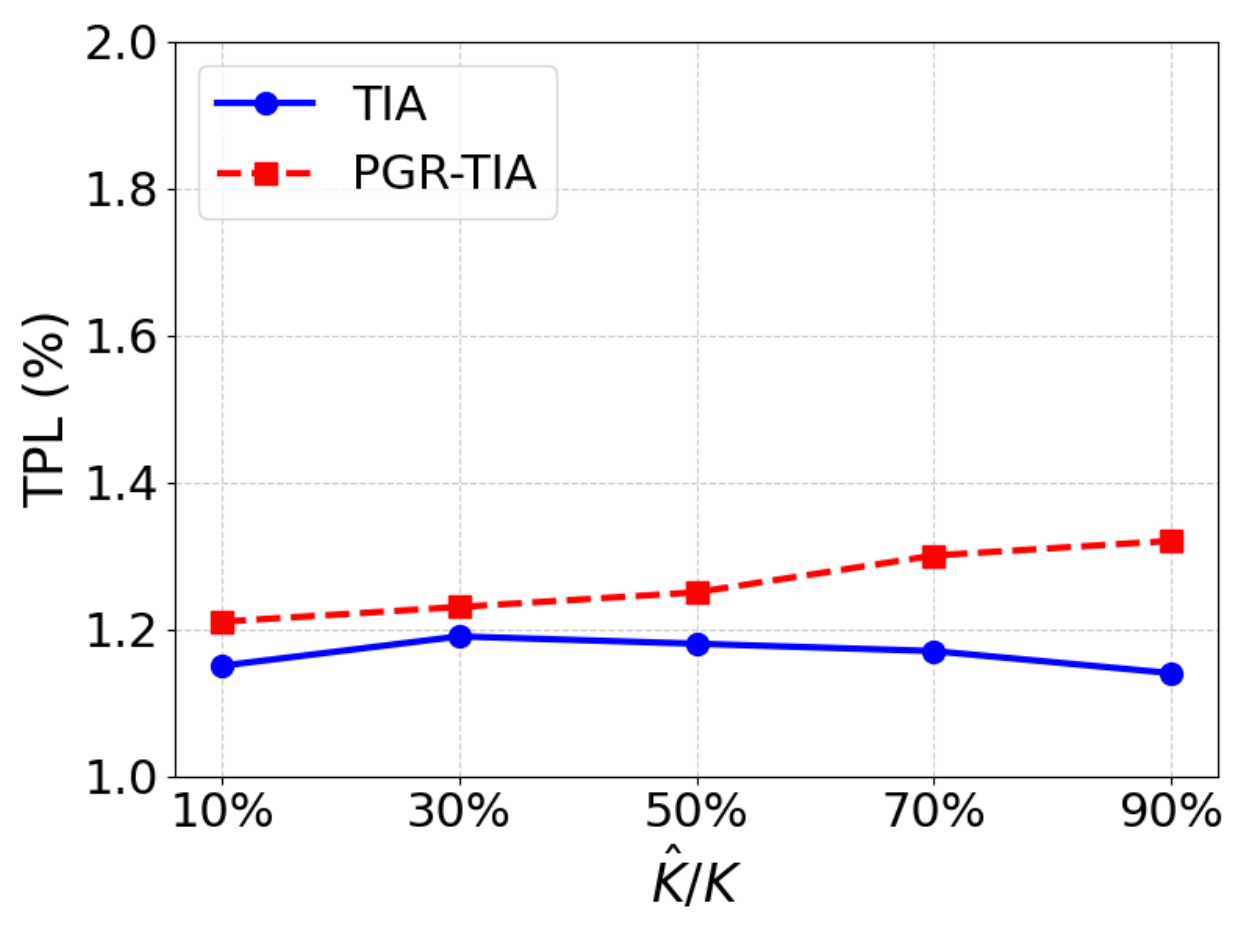} 
        \caption{Citeseer}
        \label{fig:PGR_TIA_citeseer}
    \end{subfigure}
     \caption{Topology privacy leakage (\%) of \system\ against both TIA and PGR-TIA.} 
        \label{fig:robustness}
\end{figure}

\begin{figure*}[h]
	\begin{center}	\includegraphics[width=0.85\linewidth]{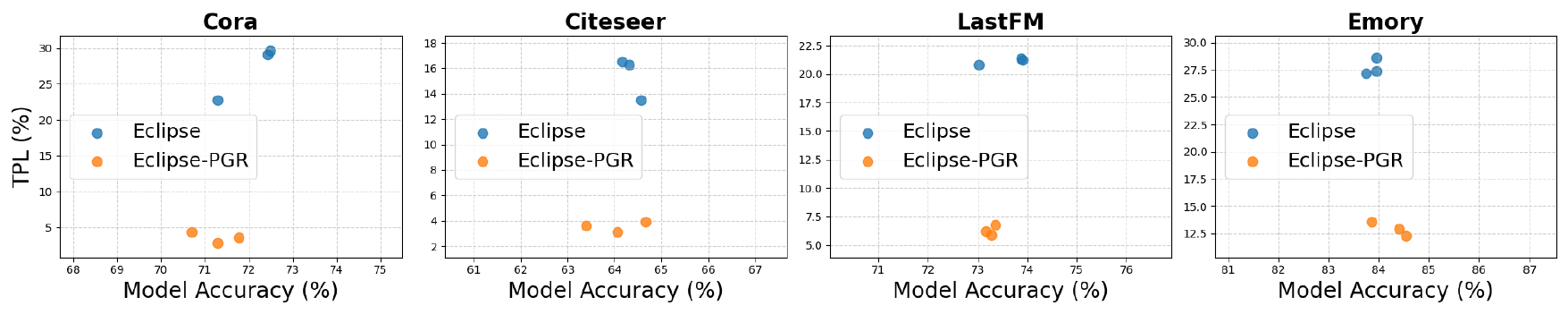}
     \vspace{-0.1in}
		\caption{Trade-off between TPL and model accuracy of DP-PGR, with Eclipse~\cite{tang2024edge} as the edge-DP approach}
        \label{figure:DP-PGR}
	\end{center}
\end{figure*}

\subsection{Robustness of \system\ against PGR-TIA ($\mathbf{RQ_4}$)}
\label{sc:pgr-tia-performance}

To assess the robustness of \system\ against PGR-TIA, we compare its topology privacy leakage (TPL) under both the original TIAs and their corresponding PGR-TIA variants. Recall that the filtering capability of PGR-TIA depends on the value of $\hat{K}$ (Section~\ref{sc:grey-box}), which represents the number of edges in the reconstructed graph. Accordingly, we vary $\hat{K}$ from 10\% to 90\% of the original graph’s size to evaluate different attack strengths. 
As shown in Figure~\ref{fig:robustness}, \system\ experiences only a marginal increase in TPL - no more than 0.4\% - when subjected to PGR-TIA compared to the original TIAs, across all settings. 
These results highlight the robustness of \system\ in preserving topological privacy, even under more informed and adaptive attacks.

\subsection{Performance of DP-PGR  ($\mathbf{RQ_5}$)}

To evaluate the performance of DP-PGR, 
we deploy Eclipse \cite{tang2024edge} on top of \system\ (named Eclipse-PGR), using the privacy budget $\epsilon = \{1, 5, 9\}$.  
The results of DP-PGR based on other edge-DP approaches can be found in Appendix~\ref{appendix:DP-PGR}.  

Figure~\ref{figure:DP-PGR} presents the TPL and model accuracy results for both Eclipse and Eclipse-PGR. First, with the addition of edge-level DP protection, Eclipse-PGR achieves significantly better topology privacy than PGR, showing a substantial reduction in TPL across all four datasets. Second, Eclipse-PGR maintains model accuracy comparable to that of PGR in all settings.
These results demonstrate that \system\ can be seamlessly integrated with existing edge-level DP methods to enhance topology privacy protection without compromising model performance.





\section{Discussions}
\label{sc:discussion}

\subsection{Scalability of TIA and \system}

So far, our evaluation of TIAs and \system\ has focused on small subgraphs. But how do these methods perform when applied to GNN models trained on full, large-scale graphs? To answer this, we consider the entire graph as the attack target and assess the performance of both TIAs and \system.
As shown in Table~\ref{table:attack on whole graph}, both TIAs and \system\ remain highly effective at scale. Specifically, I-TIA continues to achieve a TPL of 100\%, while M-TIA and C-TIA yield TPLs ranging from 6.22\% to 11.08\%. In contrast, \system\ successfully reduces TPL for I-TIA to just 0.04\%–0.13\%, and brings M-TIA and C-TIA TPLs below 4\% in all cases.

However, for extremely large graphs, both TIA and \system\ may require substantial computational resources, which may not always be feasible. To address this scalability challenge, the graph can be partitioned into smaller subgraphs, allowing TIAs to be executed independently on each part. The inferred subgraphs can then be merged into a unified graph structure.
\system\ can adopt a similar strategy: it constructs multiple synthetic subgraphs $\hat{G}_1, \dots, \hat{G}_t$, each with an edge set disjoint from the original graph 
$\oldgraph$. These subgraphs are then stitched together to form a global synthetic graph $\newgraph=\cup_{i=1}^{t}\hat{G}_i$, on which a new GNN model  $\newmodel$ is trained.

\begin{table}[t!]
\caption{Performance of TIAs and \system\ on large graphs}
\vspace{-0.1in}
\label{table:attack on whole graph}
\centering
\small
\begin{tabular}{c|ccc|ccc}
\hline
\multirow{2}{*}{\textbf{Dataset}} & \multicolumn{3}{c|}{\textbf{Without defense}}                                               & \multicolumn{3}{c}{\textbf{PGR}}                                                           \\ \cline{2-7} 
                                  & \multicolumn{1}{c|}{\textbf{M-TIA}} & \multicolumn{1}{c|}{\textbf{C-TIA}} & \textbf{I-TIA} & \multicolumn{1}{c|}{\textbf{M-TIA}} & \multicolumn{1}{c|}{\textbf{C-TIA}} & \textbf{I-TIA} \\ \hline
Cora                              & \multicolumn{1}{c|}{10.8}           & \multicolumn{1}{c|}{8.65}           & 100.00            & \multicolumn{1}{c|}{0.91}           & \multicolumn{1}{c|}{0.27}           & 0.13           \\ \hline
Citeseer                          & \multicolumn{1}{c|}{11.08}          & \multicolumn{1}{c|}{9.76}           & 100.00            & \multicolumn{1}{c|}{0.57}           & \multicolumn{1}{c|}{0.26}           & 0.06           \\ \hline
Emory                             & \multicolumn{1}{c|}{7.06}           & \multicolumn{1}{c|}{6.22}           & 100.00               & \multicolumn{1}{c|}{3.91}           & \multicolumn{1}{c|}{3.62}           &    0.04            \\ \hline
\end{tabular}
\end{table}

\subsection{From Topology Inference to Graph Property Inference} 

\begin{table}[]
\small
\caption{Accuracy of graph property inference by TIA (GCN model, Cora dataset). The accuracy of community detection is measured using the Adjusted Rand Index (ARI), and the accuracy of influential node detection is assessed using the F1-score.}
\label{tab:Topology Privacy Leakage}
\begin{tabular}{c|c|c|c}
\hline
\multirow{2}{*}{\textbf{Attack}} & {\bf Community}  & \multicolumn{2}{c}{{\bf Detection of influential nodes}} \\\cline{3-4} 
& {\bf detection} & {\bf Degree-based} & {\bf Centrality-based} \\\hline
M-TIA           & 0.15                                                                   & 0.16                                                                               & 0.18                                                                                   \\ \hline
C-TIA           & 0.13                                                                  & 0.13                                                                              & 0.19                                                                                  \\ \hline
I-TIA           & 1.00                                                                     & 1.00                                                                                & 1.00                                                                                    \\ \hline
\end{tabular}
\end{table}

Since the topology of graphs can reveal critical structural information, an important question arises: can TIAs also expose key properties of the training graph? To investigate this, we evaluate the effectiveness of TIAs in inferring two fundamental graph properties: {\em community structures} and {\em influential nodes}.

\begin{ditemize}
    \item {\bf Community detection:} We use Louvain~\cite{blondel2008fast}, a SOTA community detection algorithm, on both the private graph and the attack graph. We use the Adjusted Rand Index (ARI)~\cite{gates2016monte} (ranging in [-1, 1]) to measure the similarity between the communities detected in each graph. A higher ARI indicates a higher accuracy of community detection from the attack graph.
    
    \item {\bf Influential nodes:} We define {\em influential nodes} as the top 100 nodes ranked by degree and by centrality, respectively. We measure the precision and recall of the influential nodes identified in the attack graph, and report the F1-score computed from the precision and recall.  
\end{ditemize} 

Table~\ref{tab:Topology Privacy Leakage} presents the results for the GCN model trained on the Cora dataset. All three TIAs successfully infer communities and influential nodes from the model, with I-TIA achieving the highest accuracy in both tasks. These results demonstrate that TIAs, particularly I-TIA, can reveal community structures and influential nodes in the private graph, exposing privacy risks that extend beyond simple link disclosure.

\subsection{Mitigation of Edge-level Privacy Vulnerability by \system} 

\begin{table}[t!]
\centering
\footnotesize
\caption{Mitigation of edge-level privacy vulnerability by \system. The accuracy (in \%) of link membership inference attacks is reported. }
\vspace{-0.1in}
\label{table:edge privacy}
\begin{tabular}{c|c|cc|cc}
\hline
\multirow{3}{*}{\textbf{Model}} & \multirow{3}{*}{\textbf{Dataset}} & \multicolumn{2}{c|}{\multirow{2}{*}{\textbf{Without defense}}}       & \multicolumn{2}{c}{\multirow{2}{*}{\textbf{PGR}}}                   \\
                                &                                   & \multicolumn{2}{c|}{}                                               & \multicolumn{2}{c}{}                                                \\ \cline{3-6} 
                                &                                   & \multicolumn{1}{c|}{\textbf{StealLink}} & \textbf{Infiltrator} & \multicolumn{1}{c|}{\textbf{StealLink}} & \textbf{Infiltrator} \\ \hline
\multirow{3}{*}{GCN}            & Cora                              & \multicolumn{1}{c|}{84.86}              & 100.00                        & \multicolumn{1}{c|}{66.10}               & 49.74                     \\ \cline{2-6} 
                                & Citeseer                          & \multicolumn{1}{c|}{88.12}              & 100.00                       & \multicolumn{1}{c|}{61.17}              & 50.98                     \\ \cline{2-6} 
                                & Emory                             & \multicolumn{1}{c|}{62.93}              & 100.00                        & \multicolumn{1}{c|}{50.54}              & 49.20                      \\ \hline
\multirow{3}{*}{GAT}            & Cora                              & \multicolumn{1}{c|}{85.16}              & 91.72                     & \multicolumn{1}{c|}{63.97}              & 49.74                     \\ \cline{2-6} 
                                & Citeseer                          & \multicolumn{1}{c|}{86.81}              & 98.62                     & \multicolumn{1}{c|}{63.23}              & 50.98                     \\ \cline{2-6} 
                                & Emory                             & \multicolumn{1}{c|}{60.62}              & 74.05                     & \multicolumn{1}{c|}{51.21}              & 50.12                     \\ \hline
\multirow{3}{*}{GraphSAGE}      & Cora                              & \multicolumn{1}{c|}{81.15}              & 97.02                     & \multicolumn{1}{c|}{65.98}              & 49.74                     \\ \cline{2-6} 
                                & Citeseer                          & \multicolumn{1}{c|}{90.30}               & 100.00                        & \multicolumn{1}{c|}{61.17}              & 50.98                     \\ \cline{2-6} 
                                & Emory                             & \multicolumn{1}{c|}{63.86}              & 78.44                     & \multicolumn{1}{c|}{50.98}              & 49.98                     \\ \hline
\end{tabular}
\end{table}

We have demonstrated that \system\ effectively mitigates topology-level privacy leakage in GNNs. But can it also defend against edge-level privacy attacks? To answer this question, we evaluate \system's resilience against the {\em link membership inference attacks} (LMIAs). Specifically, we consider two state-of-the-art LMIA methods: {\em StealLink}~\cite{he2021stealing} and {\em Infiltrator}~\cite{meng2023devil}. To evaluate the accuracy of these attacks, we randomly sample 500 connected node pairs (members) and 500 unconnected node pairs (non-members) to form the test set. We measure attack accuracy as the proportion of correctly identified member and non-member edges. We repeat over five independent trials, and report the average results.

Table~\ref{table:edge privacy} shows the attack accuracy of StealLink and Infiltrator against the original GNN model and the model protected by \system. The results reveal a clear trend: \system\ significantly reduces the effectiveness of both attacks, with Infiltrator’s accuracy dropping to nearly 50\% — equivalent to random guessing. These findings highlight that \system\ not only strengthens topology-level privacy but also provides strong protection against edge-level membership inference threats.

\subsection{Alternative Topology Privacy Evaluation Metric} 
\label{sc:alternativemetric}
TPL is primarily measured by the Jaccard similarity between the original and reconstructed graphs. Are there alternative metrics for evaluating topology privacy leakage? 
To answer this question, we design the following metric that assess topology privacy leakage through the {\em precision} and {\em recall} of the edges in the attack graph. Specifically, let $E_T$ and $E_{A}$ be the edges in the private graph and the attack graph, respectively. The precision is measured as $\frac{|E_T \cap E_A|}{|E_A|}$, and recall is measured as $\frac{|E_T \cap E_A|}{|E_T|}$. Finally, we compute the F1-score from the precision and recall. 

Table~\ref{tab:F1-score Metric} presents the TPL results, measured by Jaccard similarity and F1-score, before and after deploying \system, with GCN trained on the Cora dataset as the target model. Both metrics yield consistent results across all settings, showing that I-TIA causes the highest privacy leakage, while C-TIA results in the lowest.

\begin{table}[]
\small
\caption{Topology privacy leakage of TIAs measured by Jaccard similarity (Sim.) and F1-score (GCN model, Cora dataset).}
\label{tab:F1-score Metric}
\begin{tabular}{c|cc|cc|cc}
\hline
\multirow{2}{*}{\textbf{Method}} & \multicolumn{2}{c|}{\textbf{M-TIA}}                   & \multicolumn{2}{c|}{\textbf{C-TIA}}                   & \multicolumn{2}{c}{\textbf{I-TIA}}                    \\ \cline{2-7} 
                                 & \multicolumn{1}{c|}{\textbf{F1-score}} & \textbf{Sim.} & \multicolumn{1}{c|}{\textbf{F1-score}} & \textbf{Sim.} & \multicolumn{1}{c|}{\textbf{F1-score}} & \textbf{Sim.} \\ \hline
W/O \system                      & \multicolumn{1}{c|}{42.12}             & 28.10         & \multicolumn{1}{c|}{37.69}             & 25.53        & \multicolumn{1}{c|}{100.00}               & 100.00          \\ \hline
With \system                     & \multicolumn{1}{c|}{6.20}               & 3.01         & \multicolumn{1}{c|}{5.34}              & 2.70          & \multicolumn{1}{c|}{1.89}              & 0.81         \\ \hline
\end{tabular}
\end{table}

\subsection{Relaxation of Non-overlapping Constraint of \system}

\begin{figure}[t!]
    \centering
    \begin{subfigure}[b]{0.23\textwidth} 
        \includegraphics[width=\textwidth]{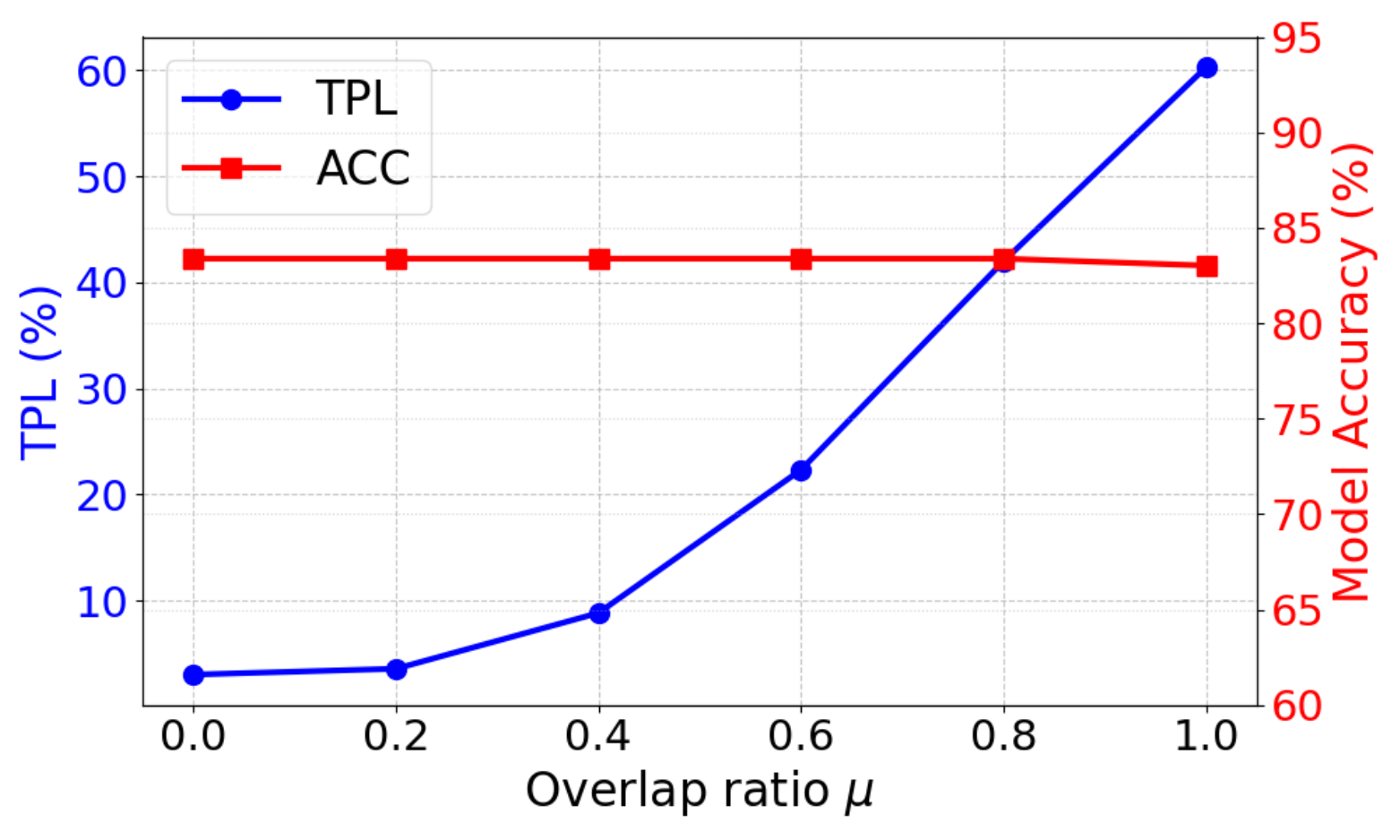} 
        \caption{Cora}
        \label{fig:impact_of_mu_cora}
    \end{subfigure}
    \begin{subfigure}[b]{0.23\textwidth}
        \includegraphics[width=\textwidth]{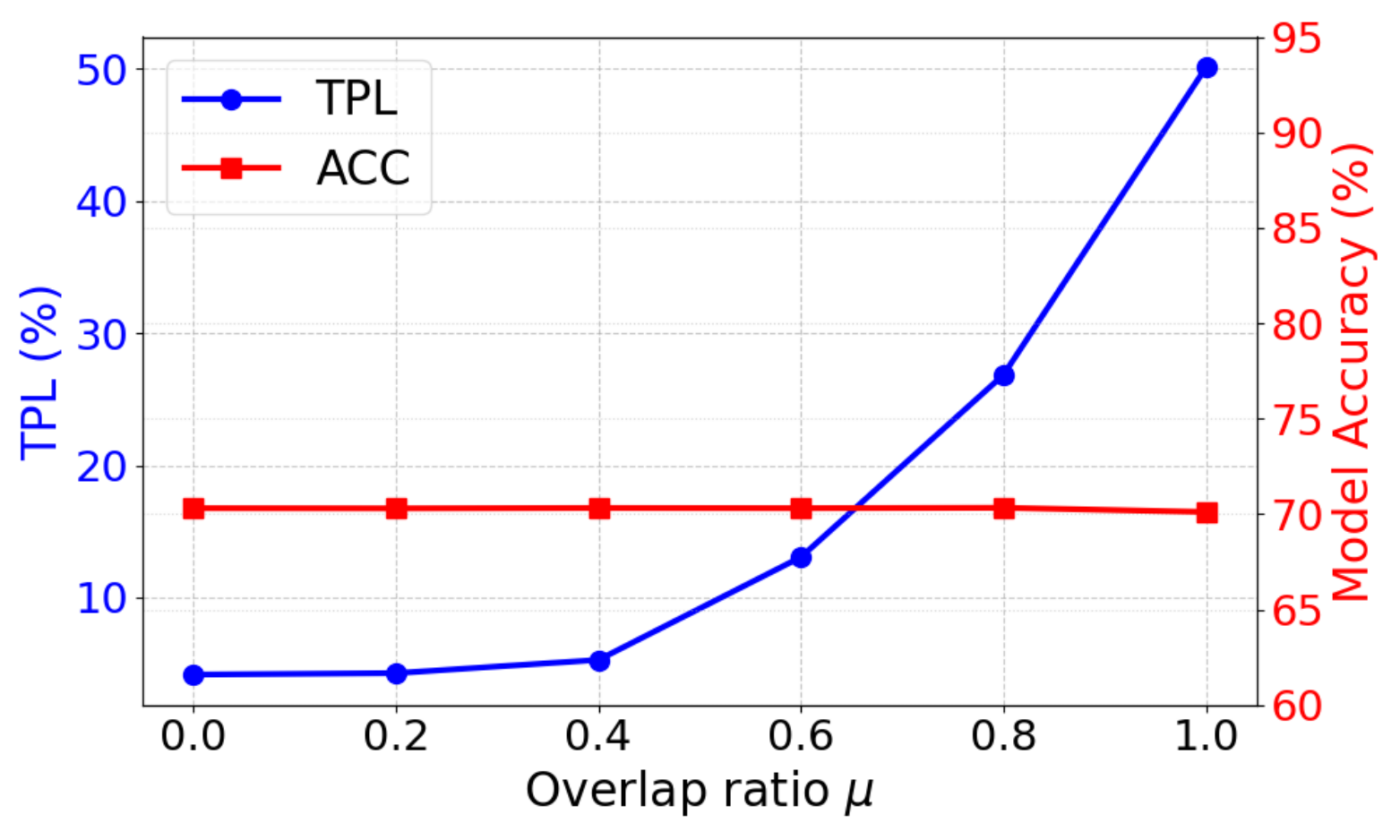} 
        \caption{Citeseer}
        \label{fig:impact_of_mu_citeseer}
    \end{subfigure}
      \vspace{-0.1in}
		\caption{Impact of overlapping ratio $\mu$ on \system's performance. The left (blue) and the right (red) y-axes indicate topology privacy leakage and model accuracy, respectively.}
         \label{figure:impact_C}
         \vspace{-0.1in}
\end{figure}

In the paper, we primarily focus on constructing a graph whose edges do not overlap with those in the original training graph. However, such a non-overlapping graph may not exist when $\oldgraph$ is extremely dense - for instance, in the case of a fully connected graph. To address this limitation, we relax the strict non-overlapping constraint by allowing a controlled level of overlap, defined as $\frac{|E \cap \hat{E}|}{|\hat{E}|} \leq \mu$, where $\mu \in [0, 1]$ is a user-defined parameter. This relaxed constraint can be easily incorporated into \system\ by adding a simple check of the overlapping degree between $E$ and $|\hat{E}|$ in Algorithm~\ref{algorithm1}.

We evaluate the impact of the overlap constraint on \system's performance by varying the overlap ratio $\mu$ from 0 to 1. We measure both the TPL and the model accuracy for these settings and presents the results on the Cora and Citeseer datasets in Figure~\ref{figure:impact_C}. As $\mu$ increases, we observe a steady rise in TPL, indicating weaker privacy protection.  However, the model accuracy remains largely unaffected, demonstrating \system's robustness to edge overlap.

Notably, it is not easy to equip PGR-TIA with the overlapping constraint, as the edges in the inferred graph $\hat{G}$ may still exist in the original graph, making the adversary  difficult to refine the candidate edge set for the second round of TIA. We will explore this scenario in the future work. 

\section{Conclusion}
\label{sec:conclusion}
In this paper, we addressed the critical challenge of preserving topology privacy in GNNs. We first introduced a suite of black-box Topology Inference Attacks (TIAs), and showed that SOTA edge-DP methods are insufficient to defend against such attacks. To tackle this issue, we proposed \system, a novel defense framework that effectively mitigates topology privacy leakage while maintaining high model accuracy. Extensive experiments across three widely-used GNN architectures and four real-world datasets demonstrate the strong protection and utility trade-off achieved by \system.  

For future work, we plan to investigate the privacy risks associated with other graph-level statistics, such as minimum and maximum node degrees, and their implications for topology inference. Additionally, we will explore new forms of adaptive attacks and evaluate the robustness of \system\ against these more sophisticated threat models. 

\nop{
\begin{acks}
This work was initiated when Jie Fu was studying at East China Normal University with Prof. Zhili Chen.
\end{acks}
}



\section*{Acknowledgments}
We thank the reviewers for their feedback. This work was supported by the National Science Foundation (CNS-2029038; CNS-2135988; CNS-2302689; CNS-2308730; CNS-2432533). Any opinions, findings, conclusions, or recommendations expressed in this paper are those of the authors and do not necessarily reflect the views of the funding agency. This work was initiated when Jie Fu was studying at East China Normal University with Prof. Zhili Chen.

\bibliographystyle{ACM-Reference-Format}
\balance

\appendix
\section*{Appendix}

\section{TPL of Random Guessing}
\label{Appendix:randomguess}

\begin{table}[h!]
\small 
\caption{TPL (\%) of random guessing}
\label{tab:R-TIA}
\begin{tabular}{l|c|c|c|c}
\hline
                           & \textbf{Cora} & \textbf{Citeseer} & \textbf{LastFM} & \textbf{Emory}  \\ \hline
\multicolumn{1}{c|}{Random guessing} & 1.66          & 1.64              & 4.27            & 10.09              \\ \hline
\end{tabular}
\end{table}

Given a graph, let $N$ and $K$ be its number of nodes and edges, respectively, we randomly choose $K$ node pairs from all possible pairs and construct edges between these pairs, producing the inferred graph $G_{A}$. We repeat this process for ten times and report the average TPL of all the trials in Table~\ref{tab:R-TIA}.

\section{Choice of Privacy Budget Value for Edge-DP Approaches} 
\label{appendix:C1}

\begin{table*}[t!]
\footnotesize
\caption{TPL and model accuracy (ACC) of six edge-DP approaches under various privacy budgets ($\epsilon$).}
\centering
\begin{tabular}{c|c|cc|cc|cc|cc}
\hline
\multirow{2}{*}{\textbf{Edge-DP approach}} & \multirow{2}{*}{\textbf{$\epsilon$}} & \multicolumn{2}{c|}{\textbf{Cora}} & \multicolumn{2}{c|}{\textbf{Citeseer}} & \multicolumn{2}{c|}{\textbf{LastFM}} & \multicolumn{2}{c}{\textbf{Emory}} \\ \cline{3-10} 
                                           &                               & \multicolumn{1}{c|}{\textbf{ACC}} & \textbf{TPL} & \multicolumn{1}{c|}{\textbf{ACC}} & \textbf{TPL} & \multicolumn{1}{c|}{\textbf{ACC}} & \textbf{TPL} & \multicolumn{1}{c|}{\textbf{ACC}} & \textbf{TPL} \\ \hline
\multirow{5}{*}{Eclipse}                   & 1                             & \multicolumn{1}{c|}{71.28}        & 22.73        & \multicolumn{1}{c|}{64.56}        & 13.49        & \multicolumn{1}{c|}{73.02}        & 20.81        & \multicolumn{1}{c|}{83.95}        & 27.42        \\
                                           & 3                             & \multicolumn{1}{c|}{72.01}        & 23.90         & \multicolumn{1}{c|}{63.99}        & 16.14        & \multicolumn{1}{c|}{73.28}        & 20.91        & \multicolumn{1}{c|}{83.82}        & 28.21        \\
                                           & 5                             & \multicolumn{1}{c|}{72.43}        & 29.05        & \multicolumn{1}{c|}{64.31}        & 16.26        & \multicolumn{1}{c|}{73.87}        & 21.34        & \multicolumn{1}{c|}{83.95}        & 28.62        \\
                                           & 7                             & \multicolumn{1}{c|}{72.45}        & 29.50         & \multicolumn{1}{c|}{63.99}        & 16.51        & \multicolumn{1}{c|}{73.87}        & 20.94        & \multicolumn{1}{c|}{83.78}        & 27.36        \\
                                           & 9                             & \multicolumn{1}{c|}{72.48}        & 29.68        & \multicolumn{1}{c|}{64.16}        & 16.54        & \multicolumn{1}{c|}{73.91}        & 21.22        & \multicolumn{1}{c|}{83.74}        & 27.16        \\ \hline
\multirow{5}{*}{LPGNet}                    & 1                             & \multicolumn{1}{c|}{63.15}        & 10.82        & \multicolumn{1}{c|}{64.43}        & 2.08         & \multicolumn{1}{c|}{46.31}        & 22.54        & \multicolumn{1}{c|}{70.36}        & 18.31        \\
                                           & 3                             & \multicolumn{1}{c|}{69.80}         & 21.99        & \multicolumn{1}{c|}{67.84}        & 2.11         & \multicolumn{1}{c|}{54.69}        & 22.46        & \multicolumn{1}{c|}{71.16}        & 19.51        \\
                                           & 5                             & \multicolumn{1}{c|}{69.96}        & 22.74        & \multicolumn{1}{c|}{68.21}        & 2.65         & \multicolumn{1}{c|}{56.52}        & 22.68        & \multicolumn{1}{c|}{71.36}        & 19.57        \\
                                           & 7                             & \multicolumn{1}{c|}{70.13}        & 24.67        & \multicolumn{1}{c|}{68.57}        & 3.53         & \multicolumn{1}{c|}{56.82}        & 22.74        & \multicolumn{1}{c|}{71.38}        & 19.38        \\
                                           & 9                             & \multicolumn{1}{c|}{70.19}        & 25.32        & \multicolumn{1}{c|}{68.64}        & 4.83         & \multicolumn{1}{c|}{56.99}        & 22.83        & \multicolumn{1}{c|}{71.38}        & 19.48        \\ \hline
\multirow{5}{*}{PrivGraph}                 & 1                             & \multicolumn{1}{c|}{42.43}        & 2.95         & \multicolumn{1}{c|}{43.62}        & 3.13         & \multicolumn{1}{c|}{45.01}        & 5.58         & \multicolumn{1}{c|}{38.05}        & 14.35        \\
                                           & 3                             & \multicolumn{1}{c|}{42.69}        & 10.96        & \multicolumn{1}{c|}{44.29}        & 3.36         & \multicolumn{1}{c|}{45.12}        & 8.25         & \multicolumn{1}{c|}{45.31}        & 16.64        \\
                                           & 5                             & \multicolumn{1}{c|}{43.86}        & 13.23        & \multicolumn{1}{c|}{44.76}        & 3.51         & \multicolumn{1}{c|}{46.79}        & 19.11        & \multicolumn{1}{c|}{50.29}        & 16.91        \\
                                           & 7                             & \multicolumn{1}{c|}{43.88}        & 13.38        & \multicolumn{1}{c|}{44.86}        & 3.56         & \multicolumn{1}{c|}{46.89}        & 19.21        & \multicolumn{1}{c|}{50.43}        & 17.06        \\
                                           & 9                             & \multicolumn{1}{c|}{43.97}        & 16.77        & \multicolumn{1}{c|}{43.92}        & 3.39         & \multicolumn{1}{c|}{49.92}        & 21.74        & \multicolumn{1}{c|}{49.28}        & 17.12        \\ \hline
\multirow{5}{*}{GAP}                       & 1                             & \multicolumn{1}{c|}{52.15}        & 2.84         & \multicolumn{1}{c|}{56.24}        & 2.22         & \multicolumn{1}{c|}{62.35}        & 10.74        & \multicolumn{1}{c|}{63.99}        & 13.11        \\
                                           & 3                             & \multicolumn{1}{c|}{54.37}        & 3.32         & \multicolumn{1}{c|}{57.91}        & 2.65         & \multicolumn{1}{c|}{64.72}        & 16.49        & \multicolumn{1}{c|}{67.10}         & 13.61        \\
                                           & 5                             & \multicolumn{1}{c|}{57.61}        & 3.78         & \multicolumn{1}{c|}{60.92}        & 3.09         & \multicolumn{1}{c|}{66.38}        & 16.87        & \multicolumn{1}{c|}{67.71}        & 13.89        \\
                                           & 7                             & \multicolumn{1}{c|}{59.45}        & 4.63         & \multicolumn{1}{c|}{62.72}        & 3.41         & \multicolumn{1}{c|}{67.55}        & 16.53        & \multicolumn{1}{c|}{67.97}        & 13.91        \\
                                           & 9                             & \multicolumn{1}{c|}{61.39}        & 4.79         & \multicolumn{1}{c|}{63.71}        & 4.09         & \multicolumn{1}{c|}{68.19}        & 16.34        & \multicolumn{1}{c|}{68.21}        & 13.97        \\ \hline
\multirow{5}{*}{LapEdge}                   & 1                             & \multicolumn{1}{c|}{23.76}        & 2.43         & \multicolumn{1}{c|}{23.95}        & 2.59         & \multicolumn{1}{c|}{16.25}        & 5.90          & \multicolumn{1}{c|}{27.19}        & 12.54        \\
                                           & 3                             & \multicolumn{1}{c|}{24.42}        & 2.98         & \multicolumn{1}{c|}{25.18}        & 2.77         & \multicolumn{1}{c|}{17.12}        & 6.34         & \multicolumn{1}{c|}{50.39}        & 30.55        \\
                                           & 5                             & \multicolumn{1}{c|}{30.34}        & 17.53        & \multicolumn{1}{c|}{26.62}        & 13.31        & \multicolumn{1}{c|}{26.37}        & 17.89        & \multicolumn{1}{c|}{80.92}        & 80.84        \\
                                           & 7                             & \multicolumn{1}{c|}{61.63}        & 72.24        & \multicolumn{1}{c|}{45.56}        & 53.59        & \multicolumn{1}{c|}{54.89}        & 58.12        & \multicolumn{1}{c|}{87.10}         & 95.30         \\
                                           & 9                             & \multicolumn{1}{c|}{77.70}         & 96.35        & \multicolumn{1}{c|}{61.69}        & 90.61        & \multicolumn{1}{c|}{72.70}         & 93.61        & \multicolumn{1}{c|}{88.73}        & 97.78        \\ \hline
\multirow{5}{*}{EdgeRand}                  & 1                             & \multicolumn{1}{c|}{29.95}        & 10.60         & \multicolumn{1}{c|}{19.84}        & 6.56         & \multicolumn{1}{c|}{20.54}        & 10.37        & \multicolumn{1}{c|}{27.19}        & 23.58        \\
                                           & 3                             & \multicolumn{1}{c|}{29.95}        & 22.59        & \multicolumn{1}{c|}{22.58}        & 29.10         & \multicolumn{1}{c|}{25.62}        & 50.16        & \multicolumn{1}{c|}{51.89}        & 66.37        \\
                                           & 5                             & \multicolumn{1}{c|}{60.42}        & 71.05        & \multicolumn{1}{c|}{46.63}        & 72.14        & \multicolumn{1}{c|}{59.47}        & 85.44        & \multicolumn{1}{c|}{83.72}        & 92.88        \\
                                           & 7                             & \multicolumn{1}{c|}{74.04}        & 96.22        & \multicolumn{1}{c|}{58.95}        & 95.25        & \multicolumn{1}{c|}{69.75}        & 97.39        & \multicolumn{1}{c|}{89.05}        & 98.98        \\
                                           & 9                             & \multicolumn{1}{c|}{81.25}        & 99.98        & \multicolumn{1}{c|}{68.88}        & 99.51        & \multicolumn{1}{c|}{79.23}        & 99.95        & \multicolumn{1}{c|}{90.31}        & 99.77        \\ \hline
\end{tabular}
\label{table:TPL and model performance}
\end{table*}

To show the trade-off between privacy and model accuracy of edge-DP approaches, we measure both topology privacy leakage and model accuracy (ACC) of the six SOTA edge-DP approaches under various privacy budget $\epsilon \in \{1,3,5,7,9\}$. 
Table~\ref{table:TPL and model performance} reports the model accuracy and TPL results across the four datasets. 
Among all the settings, $\epsilon=7$ yields the best trade-off between model accuracy and privacy protection. This supports us to choose  $\epsilon=7$. 
\section{Proof of Theorem \ref{theorem:DP-PGR}}
\label{appendix:proofdp}

Following Algorithm~\ref{algorithm1}, \system\ requires access to the private graph $\oldgraph$ at two stages: {\bf (i) Model training}:  The model $\oldmodel$ is trained on $\oldgraph$ to generate the output $Y_P$ (Line 2); {\bf (ii) Graph overlap constraint}: The overlap between the generated graph $\newgraph$ and $\oldgraph$ has to be evaluated at each iteration (Line 8). The challenge is to ensure edge-level DP guarantees for these two direct accesses to the private graph.
For the model training phase, edge-DP methods can ensure that the output $Y_p$ satisfies $(\epsilon,\delta)$-edge DP. For the graph overlap constraint phase, we choose to remove the constraint to avoid accessing the original graph. This allows us to leverage the post-processing property~\cite{dwork2014algorithmic} of differential privacy to guarantee that DP-PGR satisfies $(\epsilon,\delta)$-edge DP.

\section{Additional Experimental Results}

\subsection{Impact of Number of GNN layers on
 Performance of \system.}
\label{appendix:numberoflayers}
\begin{figure*}[h]
	\begin{center}
\includegraphics[width=0.9\linewidth]{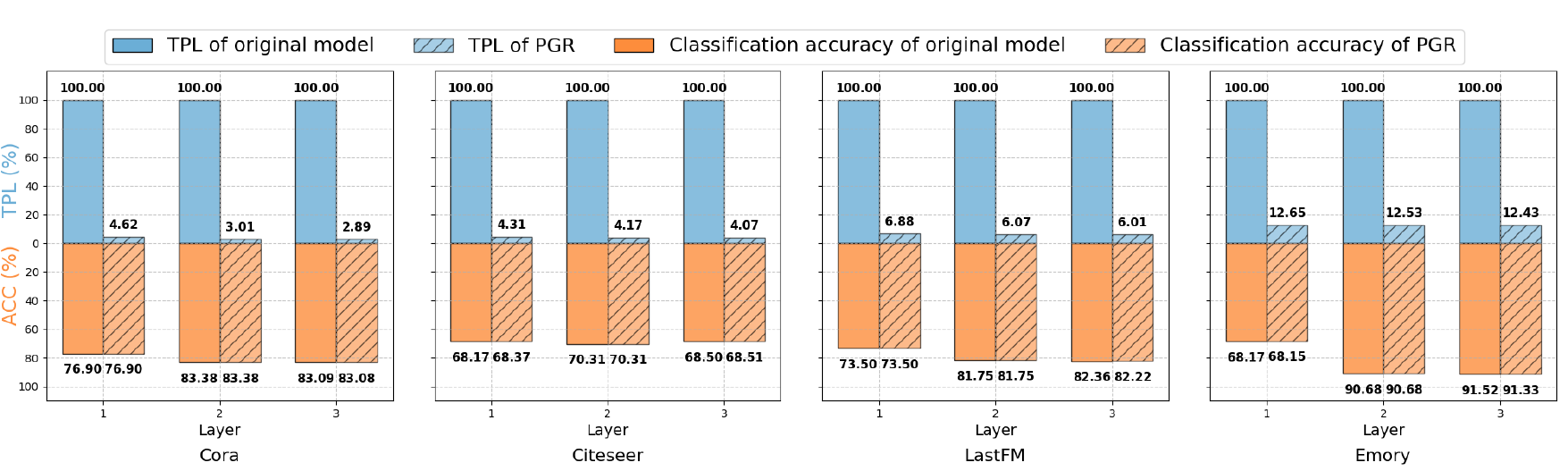}
		\caption{Impact of the number of GNN layers on both topology privacy leakage (TPL) and model accuracy. The x-axis indicates the number of GCN layers. The y-axis denotes both TPL (in blue color) and model accuracy (in orange color).}
        \label{figure:impact_layer}
        \vspace{-0.2in}
	\end{center}
\end{figure*}

To evaluate the impact of the number of GNN layers on \system's performance, we set up three GCNs with 1, 2, and 3 layers, respectively. There are 64 neurons at the first layer, 32 neurons at the second layer, and the number of neurons at the last layer equals the number of classes. We then measure both the TPL and model accuracy of these GNNs before and after deploying \system.
As shown in Figure~\ref{figure:impact_layer},  the number of GNN layers has minimal effect on both model accuracy and TPL. Specifically, the model accuracy of \system\ remains consistently close to that of the original model, while TPL stays stable as the number of GCN layers increases, across all datasets.
These findings demonstrate that \system\ is robust and can be effectively deployed on GNNs of varying complexities without compromising either model performance or the protection of topology privacy.

\begin{figure*}[h!]
	\begin{center}	\includegraphics[width=0.9\linewidth]{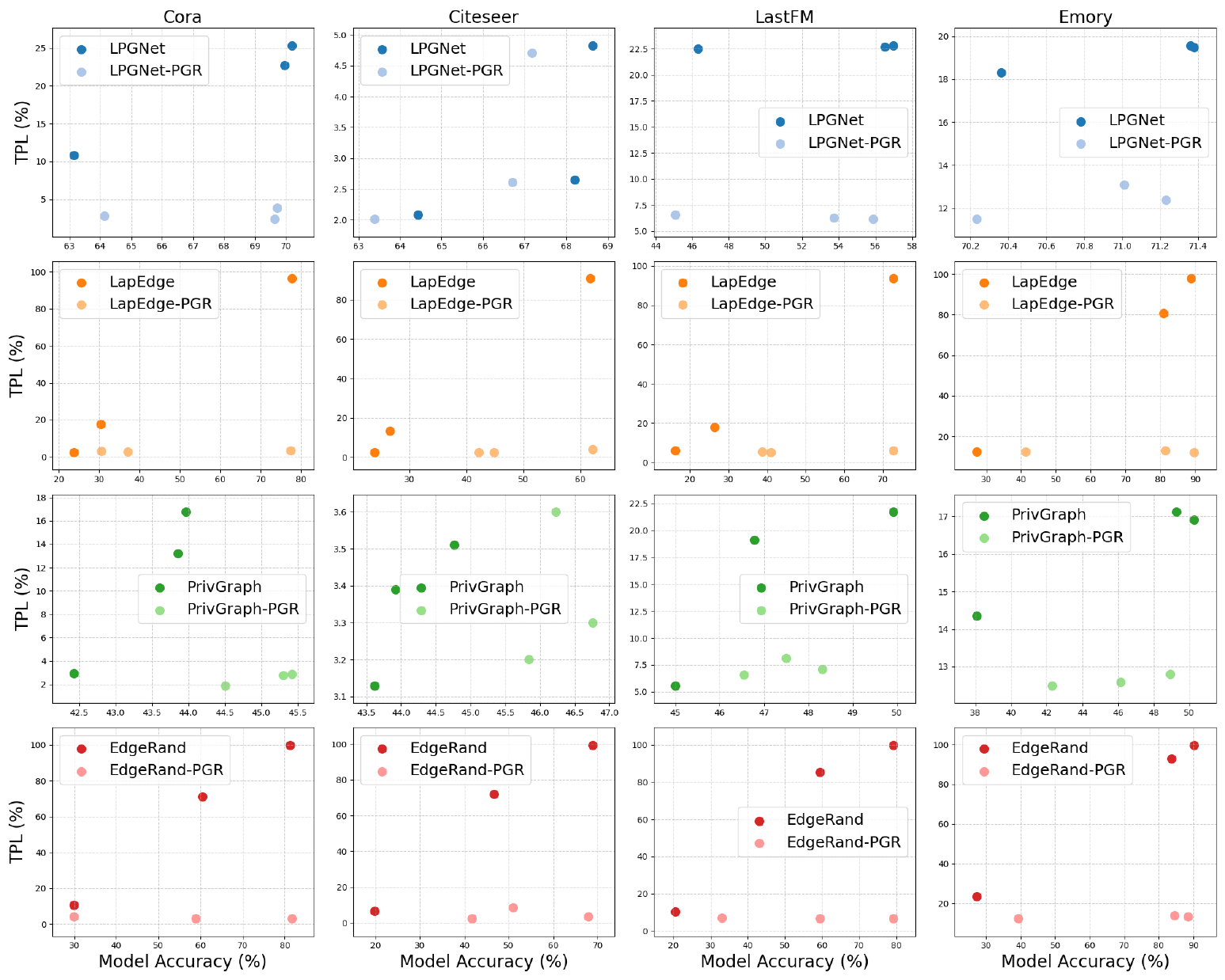}
		\caption{The trade-off between TPL and model accuracy of DP-PGR adapted from four edge-DP approaches (LPGNet~\cite{kolluri2022lpgnet}, PrivGraph~\cite{yuan2023privgraph}, LapEdge~\cite{wu2022linkteller}, and
EdgeRand~\cite{wu2022linkteller}).  The x-axis and y-axis indicate model accuracy (\%) and topology privacy leakage ($TPL$), respectively. 
  We use the privacy budget $\epsilon \in \{1,5,9\}$ for all the algorithms and datasets.}
        \label{figure:DP-PGR2}
	\end{center}
    \vspace{-0.2in}
\end{figure*}

\subsection{Trade-off between Privacy and Model Accuracy of DP-PGR} 
\label{appendix:DP-PGR}
Figure ~\ref{figure:DP-PGR2} reports the trade-off between TPL and model accuracy of the DP-PGR algorithm adapted from four edge-DP approaches (LPGNet~\cite{kolluri2022lpgnet}, PrivGraph~\cite{yuan2023privgraph}, LapEdge~\cite{wu2022linkteller}, and
EdgeRand~\cite{wu2022linkteller}). We report the best TPL result among the three TIAs for each dataset.  We use the privacy budget $\epsilon \in \{1, 5, 9\}$ for all the algorithms and measure both TPL and model accuracy for each $\epsilon$ value. 
The results show that DP-PGR can equip \system\ with DP effectively while minimizing the loss in model accuracy. This demonstrates that \system\ can be seamlessly integrated with edge-DP while addressing the trade-off between privacy protection and model accuracy. 
\clearpage

\end{document}